\documentclass[runningheads]{llncs}
\usepackage{graphicx}
\usepackage{amsmath,amssymb} 
\usepackage{color}

\usepackage{times}
\usepackage{epsfig}
\usepackage{graphicx}
\usepackage{amsmath}
\usepackage{amssymb}
\usepackage{pifont} 
\usepackage{bbm}
\usepackage{empheq}
\usepackage{rotating}
\usepackage{bm}

\usepackage{caption}
\usepackage{subcaption}
\usepackage{algorithm}
\usepackage{algorithmic}
\usepackage{multirow}
\usepackage{makecell}
\usepackage{arydshln}
\usepackage{booktabs}
\usepackage{wrapfig}

\graphicspath{ {images/} }
\DeclareMathOperator{\E}{\mathbb{E}}
\DeclareMathOperator{\R}{\mathbb{R}}

\usepackage{xspace}
\makeatletter
\DeclareRobustCommand\onedot{\futurelet\@let@token\@onedot}
\def\@onedot{\ifx\@let@token.\else.\null\fi\xspace}

\def\eg{\emph{e.g}\onedot} 
\def\ie{\emph{i.e}\onedot}

\def\etal{\emph{et al}\onedot}
\makeatother
 
\newcommand{\argmin}{\operatornamewithlimits{argmin}}
\newcommand{\argmax}{\operatornamewithlimits{argmax}}
\newcommand{\real}{\mathbb{R}}
\newcommand{\dataset}{\mathcal{D}}
\newcommand{\param}{\theta}
\newcommand{\paramdim}{P}
\newcommand{\ip}{\mathbf{x}}
\newcommand{\ipspace}{\mathcal{X}}

\newcommand{\op}{\mathbf{y}}
\newcommand{\opspace}{\mathcal{Y}}

\newcommand{\loss}{L}

\newcommand{\KL}{D_{KL}}
\newcommand{\manifold}{\mathcal{F}}

\newcommand{\numsamples}{m}
\newcommand{\bigoh}{\mathcal{O}}
\newcommand*\variant[1]{\bar{#1}}

\newcommand{\red}[1]{\textcolor{red}{#1}}
\newcommand{\pkdcomment}[1]{\textcolor{red}{\bf \small #1 --PKD}}
\newcommand{\ajcomment}[1]{\textcolor{blue}{\bf \small #1 --AJ}}

\newcommand{\SKIP}[1]{}
\newcommand*\samecontribution[1][\value{footnote}]{\footnotemark[#1]}

\newcommand{\il}{\textsc{il}}
\newcommand{\ewc}{\textsc{ewc}}
\newcommand{\pii}{\textsc{pi}}
\newcommand{\ewcp}{\textsc{ewc++}}

\usepackage[acronym,smallcaps,nowarn,section,nogroupskip,nonumberlist]{glossaries}
\glsdisablehyper{}
\newacronym{IL}{il}{incremental learning}
\newacronym{EWC}{ewc}{elastic weight consolidation}
\newacronym{PI}{pi}{path integral}
\newacronym{EWC++}{ewc++}{fast and online version of EWC}

\usepackage[pagebackref=true,breaklinks=true,letterpaper=true,colorlinks,bookmarks=false]{hyperref}

\begin{document}
\pagestyle{headings}
\mainmatter

\title{Riemannian Walk for Incremental Learning: Understanding Forgetting and Intransigence} 

\titlerunning{Riemannian Walk for Incremental Learning}

\authorrunning{Chaudhry \etal}

\author{Arslan Chaudhry\thanks{Joint first authors}, Puneet K. Dokania\samecontribution, Thalaiyasingam Ajanthan\samecontribution, Philip H. S. Torr}
\institute{
	University of Oxford, United Kingdom\\
	\email{\{firstname.lastname\}@eng.ox.ac.uk}
}

\maketitle

\begin{abstract}
Incremental learning ({\il}) has received a lot of attention
recently, however, the literature lacks a precise problem definition, proper evaluation
settings, and metrics tailored specifically for the {\il}
problem. One of the main objectives of this work is to fill these gaps so as to
provide a common ground for better understanding of {\il}. 
The main challenge for an {\il} algorithm is to update the classifier
whilst preserving existing knowledge.
We observe that, in addition to {\em forgetting}, a known issue while
preserving knowledge, {\il} also suffers from a problem we call {\em
intransigence}, inability of a model to update its knowledge. We introduce two metrics to
quantify {\em forgetting} and {\em intransigence} that allow us to understand,
analyse, and gain better insights into the behaviour of {\il}
algorithms. We present RWalk, a generalization of {\ewcp}
(our efficient version of {\ewc}~\cite{Kirkpatrick2016EWC}) and Path
Integral~\cite{Zenke2017Continual} with a theoretically grounded KL-divergence
based perspective.
We provide a thorough analysis of various {\il} algorithms
on MNIST and CIFAR-100 datasets. In these experiments, RWalk obtains superior
results in terms of accuracy, and also provides a better trade-off between
forgetting and intransigence.
\end{abstract}
\section{Introduction}
Realizing human-level intelligence requires developing systems
capable of learning new tasks continually while preserving {\em knowledge} about the old ones.
This is precisely the objective underlying incremental
learning ({\il}) algorithms. 
By definition, {\il} has ever-expanding output space, and limited or no access
to data from the previous tasks while learning a new one. This makes {\il} more challenging and fundamentally different from the classical learning paradigm where the output space is fixed and the entire dataset is available.
Recently, there have been several works in
{\il}~\cite{Kirkpatrick2016EWC,lopez2017gradient,Rebuffi16icarl,Zenke2017Continual}
with varying evaluation settings and metrics making it difficult to establish
fair comparisons.
The first objective of this work is to rectify these issues by providing
precise definitions, evaluation settings, and metrics for {\il} for the
classification task.

Let us now discuss the key points to consider while designing {\il} algorithms.
The first question is {\em `how to define knowledge: factors that quantify what
the model has learned'}.
Usually, knowledge is defined either using the input-output behaviour of the
network~\cite{hinton2015distilling,Rebuffi16icarl} or the network
parameters~\cite{Kirkpatrick2016EWC,Zenke2017Continual}.
Once the knowledge is defined, the objective then is to {\em preserve} and {\em
update} it to counteract two inherent issues with {\il} algorithms:
(1) \textit{forgetting}:
catastrophically forgetting knowledge of previous tasks; and (2)
\textit{intransigence}:
inability to update the knowledge to learn a new task.
Both of these problems require contradicting solutions and pose a trade-off for
any {\il} algorithm.

To capture this trade-off, we advocate the use of measures that evaluate
an {\il} algorithm based on its performance on the {\em past} and the {\em
present} tasks
in the hope that this will reflect in 
the algorithm's behaviour on the {\em future} unseen tasks.
Taking this into account we introduce two metrics to evaluate {\em forgetting}
and {\em intransigence}.
These metrics together with the
standard multi-class average accuracy allow us to understand, analyse, and
gain better insights into the behaviour of various {\il} algorithms.


Further, we present a generalization of two recently proposed
incremental learning algorithms, Elastic Weight Consolidation
\textsc{(ewc)}~\cite{Kirkpatrick2016EWC}, and Path Integral
\textsc{(pi)}~\cite{Zenke2017Continual}. In particular, first we show that in \ewc,
while learning a new task, the model's likelihood distribution is
regularized 
using a well known second-order approximation of the 
KL-divergence~\cite{Amari98NaturalGradient,Pascanu14NaturalGradient}, which
is equivalent to computing distance in a Riemannian manifold induced by the
Fisher Information Matrix~\cite{Amari98NaturalGradient}. To compute and update
the Fisher matrix, we use an efficient (in terms of memory) and online (in terms
of computation) approach, leading to a faster and online version of {\ewc} which
we call {\ewcp}. Note that, a similar extension to \textsc{ewc}, called
online-\textsc{ewc}, is concurrently proposed by Schwarz~\etal
~\cite{schwarz2018progress}.
Next, we modify the \textsc{pi}~\cite{Zenke2017Continual} where instead of
computing the change in the loss per unit distance in the Euclidean space
between the parameters as the measure of sensitivity, we use the approximate KL
divergence (distance in the Riemannian manifold) between the output
distributions as the distance to compute the sensitivity. This gives us the {\em
parameter importance score} which is accumulated over the optimization
trajectory effectively encoding the information about the all the seen tasks so far.
Finally, RWalk is obtained by combining {\ewcp} and the modified {\pii}. 

Furthermore, in order to counteract {\em intransigence}, 
we study different {\em sampling} strategies that store a small representative subset 
(\( \le 5\% \)) of the dataset from the previous tasks. This not only allows the
network to \textit{recall} information about the previous tasks but also helps in learning 
to \textit{discriminate} current and
previous tasks. 
Finally, we present a thorough analysis to better understand the behaviour of
{\il} algorithms on MNIST~\cite{lecun1998mnist} and
CIFAR-100~\cite{krizhevsky2009learning} datasets.
To summarize, our main contributions are:
\begin{enumerate}
  \item New evaluation metrics - {\em Forgetting} and
  {\em Intransigence} - to better understand the behaviour and performance of an
  incremental learning algorithm.
\item {\ewcp}: An efficient and online version of \textsc{ewc}.
\item RWalk: A generalization of 
{\ewcp} and {\pii} with theoretically grounded KL-divergence based
perspective providing new insights.
\item Analysis of different methods in terms of accuracy, forgetting, and intransigence.
\end{enumerate}

\vspace{-0.5cm}
\section{Problem Set-up and Preliminaries}
\label{sec:probSetup}
\vspace{-0.3cm}
Here we define the {\il} problem and discuss the practicality of two
different evaluation settings: (a) single-head; and (b) multi-head. In addition,
we review the probabilistic interpretation of neural networks and the connection
of KL-divergence with the distance in the Riemannian manifold, both of which are
crucial to our approach.

\subsection{Single-head vs Multi-head Evaluations}\label{sec:singlemulti}
\label{sec:heads}
We consider a stream of tasks, each corresponding to a set of labels.
For the $k$-th task, let $\dataset_k = \{(\ip^k_i,
y^k_i)\}_{i=1}^{n_k}$ be the dataset, where $\ip^k_i \in \ipspace$ is the input
and $y^k_i \in \op^k$ the ground truth label, and $\op^k$ is the set of
labels specific to the task. 
The main distinction between the single-head and the multi-head evaluations is
that, at test time, in {\em single-head}, the task identifier ($k$) is unknown,
whereas in {\em multi-head}, it is given. 
Therefore, for the single-head evaluation, the objective at the $k$-th
task is to learn a function $f_{\param}: \ipspace \rightarrow
\opspace^k$, where $\opspace^k=\cup_{j=1}^k \op^j$ corresponds to all the known
labels. For multi-head, as the task identifier is known, $\opspace^k= \op^k$.
For example,  
consider MNIST with $5$ tasks:
$\{\{0,1\}, \cdots, \{8,9\}\}$; trained in an incremental manner. 
Then, at the $5$-th task, for a given image, the {\em
multi-head} evaluation is to predict a class out of two labels $\{8,9\}$ for
which the $5$-th task was trained. However, the {\em single-head} evaluation at
$5$-th task is to predict a label out of all the ten classes $\{0,\cdots,
9\}$ that the model has seen thus far.

%

\subsubsection{Why single-head evaluation for {\il}?}
In the case of {\em single-head}, used by~\cite{lee2017IMM,Rebuffi16icarl},
the output space consists of all the known labels. This requires the classifier
to learn to distinguish labels from different tasks as well. 
Since, the tasks are supplied in a sequence in {\il}, while learning a new
task, the classifier must also be able to learn the inter-task discrimination with
no or limited access\footnote{Since the number of tasks are potentially
unlimited in {\il}, it is impossible to store all the previous data in a
scalable manner.} to the previous tasks' data. This is a much harder problem compared to
multi-head where the output space contains labels of the current task only.
Furthermore, single-head is a more practical setting because knowing the
subset of labels to look at a priori, which is the case in multi-head, requires extra supervisory signals, in the form of task descriptors, at test time that reduce the problem complexity. For instance, if the task
contains only one label, multi-head evaluation would be equivalent 
to knowing the ground truth label itself. 

\subsection{Probabilistic Interpretation of Neural Network Output}
\label{sec:prelim}
If the final layer of a neural network is a soft-max layer and the network is
trained using cross entropy loss, then the output may be interpreted as a
probability distribution over the categorical variables.
Thus, at a given \(\param\),
the conditional likelihood distribution learned by a neural network is  actually
a conditional multinoulli distribution defined as $p_{\param}(\op | \ip ) =
\prod_{j=1}^K p_{\param,j}^{[y=j]}$, where $p_{\param,j}$ is the soft-max
probability of the $j$-th class, $K$ are the total number of classes, $\op$ is
the one-hot encoding of length $K$, and $[\cdot]$ is Iverson bracket. A
prediction can then be obtained from the likelihood
distribution $p_{\param}(\op | \ip )$. Typically, instead of 
sampling, a label with the highest soft-max probability is chosen as the network's
prediction. Note that, if $\op$ corresponds to the ground-truth label then the
log-likelihood is exactly the same as the negative of the cross-entropy loss,
\ie, if the ground-truth corresponds to the $t$-th index of the one-hot
representation of $\op$, then  \( \log p_{\param}(\op | \ip ) = \log
p_{\param,t}\) (more details can be found in the supplementary material).

\subsection{KL-divergence as the Distance in the Riemannian
Manifold}\label{sec:kl} 
Let \( \KL(p_{\param}\| p_{\param + \Delta \param}) \) be the
KL-divergence~\cite{Kullback51KLdivergence} between the conditional likelihoods
of a neural network at \( \param \) and
\( \param + \Delta \param \), respectively. Then, assuming \( \Delta \param \to
0 \), the second-order Taylor approximation of the KL-divergence can be written
as \( \KL(p_{\param} \| p_{\param + \Delta \param}) \approx
     \frac{1}{2} \Delta \param^{\top} F_{\param} \Delta \param = \frac{1}{2} 
     \|\Delta \param\|^2_{F_{\param}} \)
     \footnote{Proof and insights are provided in the supplementary material.}, where \(
     F_{\param} \), known as the {\em empirical Fisher Information
     Matrix}~\cite{Amari98NaturalGradient,Pascanu14NaturalGradient} at $\param$,
     is defined as:
   \begin{equation}\label{eq:fisher}
F_{\param} = \E_{(\ip, \op) \sim \dataset}
\left[\left( \frac{\partial \log p_{\param}( \op | \ip) }{\partial \param} 
\right) \left( \frac{\partial \log
p_{\param}( \op | \ip) }{\partial \param} \right)^{\top}\right] \ ,
\end{equation}
where $\dataset$ is the dataset. Note that, as mentioned earlier, the
log-likelihood \( \log p_{\param}( \op | \ip) \) is the same as the negative of
the cross-entropy loss function, thus, \( F_{\param} \) can be seen as the {\em
expected loss-gradient covariance matrix}. By construction (outer product of
gradients), \( F_{\param} \) is positive semi-definite (PSD) which makes it
highly attractive for second-order optimization
techniques~\cite{Amari98NaturalGradient,Pascanu14NaturalGradient,grosse2016kronecker,LeRoux07NIPS_Topmoumoute,martens2015optimizing}.
When \( \Delta \param \to 0 \), computing KL-divergence \( \frac{1}{2} 
\|\Delta \param\|^2_{F_{\param}} \) is equivalent to computing the {\em
distance} in a Riemannian manifold\footnote{Since \( F_{\param} \) is PSD, this makes it a
pseudo-manifold.}~\cite{lee2006riemannian} induced by the Fisher information
matrix at $\param$. Since \( F_{\theta} \in \real^{\paramdim \times \paramdim} \)
and \( \paramdim \) is usually in the order of millions for neural networks, it
is practically infeasible to store \( F_{\theta} \). To handle this, similar
to~\cite{Kirkpatrick2016EWC}, we assume parameters to be independent of each
other (diagonal \( F_{\theta}\)) which results in the following approximation of
the KL-divergence:
\begin{equation}\label{eq:klapproxdiag1}
\KL(p_{\param} \| p_{\param + \Delta \param}) \approx
\frac{1}{2} \sum_{i=1}^{\paramdim} F_{\theta_i}\, \Delta \param_i^2\ ,
\end{equation}
where $\param_i$ is the $i$-th entry of $\param$. Notice, the diagonal entries of \( F_{\theta} \) are the expected square of the gradients, where the expectation is over the entire dataset. This makes \( F_{\theta} \) computation expensive as it requires a full forward-backward pass over the dataset.

\vspace{-0.4cm}
\section{Forgetting and Intransigence}\label{sec:measures}
\label{sec:eval_measures}
\vspace{-0.2cm}
Since the objective is to
continually learn new tasks while preserving knowledge about the previous ones, an {\il} algorithm
should be evaluated based on its performance both on the {\em past} and the {\em
present} tasks in the hope that this will reflect in algorithm's behaviour on the {\em
future} unseen tasks.
To achieve this, along with average accuracy, there are two crucial components
that must be quantified (1) {\em forgetting}: how much an algorithm forgets what it learned in the past;
and (2) {\em intransigence}: inability of an algorithm to learn new tasks.
Intuitively, if a model is heavily regularized over previous tasks to preserve
knowledge, it will forget less but have high intransigence. If, in
contrast, the regularization is too weak, while the intransigence will be small,
the model will suffer from catastrophic forgetting. Ideally, we want a
model that suffers less from both, thus efficiently utilizing a finite model capacity.
%
In contrast, if one observes high negative correlation between
forgetting and intransigence, which is usually the case, then, it suggests that
either the model capacity is saturated or the method does
not effectively utilize the capacity.
Before defining metrics for quantifying forgetting and intransigence, we first
define the multi-class average accuracy which will be the basis for defining
the other two metrics. Note, some other task specific measure of 
correctness (\eg, IoU for object segmentation) can also be used while the 
definitions of forgetting and intransigence remain the same.
\subsubsection{Average Accuracy ($A$)} 
Let $a_{k,j}\in[0,1]$ be the accuracy (fraction of correctly classified images)
evaluated on the held-out test set of the $j$-th task $(j \leq k)$ after
training the network incrementally from tasks $1$ to $k$. Note that, to compute \( a_{k,j}
\), the output space consists of either \( \op^j \) or \( \cup_{j=1}^k \op^j \)
depending on whether the evaluation is multi-head or single-head (refer
Sec.~\ref{sec:heads}). The average accuracy at task $k$ is then defined as \(A_k = \frac{1}{k} \sum_{j=1}^{k} a_{k,j}\).
The higher the $A_k$ the better the classifier, but this does not provide any
information about forgetting or intransigence profile of the {\il} algorithm
which would be crucial to judge its behaviour. 

\subsubsection{Forgetting Measure ($F$)}
We define forgetting for a particular task (or label) as the difference
between the {\em maximum} knowledge gained about the task throughout the 
learning process in the past and the knowledge the model currently has about it. This, in turn,
gives an estimate of how much the model forgot about the task given its current
state. Following this, for a classification problem, we quantify
forgetting for the $j$-th task after the model has been incrementally trained
up to task \( k > j \) as:
	\begin{equation}
		\label{eq:fm_msr}
		f_j^k = \max_{l \in \{1, \cdots, k-1\}} a_{l,j} - a_{k,j}\ ,\quad \forall j<k\ .
	\end{equation}
Note, $f_j^k\in[-1,1]$ is defined for $j < k$ as we are interested in
quantifying forgetting for {\em previous} tasks. 
Moreover, by normalizing against the number of tasks seen previously, the
average forgetting at $k$-th task is written as \( F_k = \frac{1}{k-1} \sum_{j=1}^{k-1} f_j^k \).
Lower $F_k$ implies less forgetting on previous tasks.
Here, instead of $\max$ one could use
 {\em expectation} or $a_{j,j}$ in order to quantify the knowledge about a task
 in the past.
However, taking $\max$ allows us to estimate forgetting along
the learning process as explained below.

\paragraph{Positive/Negative Backward Transfer ((P/N)BT):} 
Backward transfer (BT) is defined in~\cite{lopez2017gradient} as the influence
that learning a task $k$ has on the performance of a previous task ${j < k}$.
Since our objective is to measure forgetting, negative forgetting ($f_j^k < 0$)
implies positive influence on the previous task or positive backward
transfer (PBT), the opposite for NBT.
Furthermore,
in~\cite{lopez2017gradient}, $a_{j,j}$ is used in place of $\max_{l \in \{1,
\cdots, k-1\}} a_{l,j}$ (refer Eq.~\eqref{eq:fm_msr}) which makes the measure
agnostic to the {\il} process and does not effectively capture
forgetting. To understand this, let us consider an example with 4 tasks trained
in an incremental manner. We are interested in measuring forgetting of task 1
after training up to task 4. Let the accuracies be $\{a_{1,1}, a_{1,2},
a_{1,3},a_{1,4}\} = \{0.7, 0.8, 0.6, 0.5\}$. Here, forgetting measured based on
Eq.~\eqref{eq:fm_msr} is $f_1^4 = 0.3$, whereas~\cite{lopez2017gradient} would
measure it as $0.2$ (irrespective of the variations in $a_{1,2}$ and $a_{1,3}$). Hence, it does not capture the fact that there was a PBT in
the learning process. We believe, it is vital that an evaluation metric of an
{\il} algorithm considers such behaviour along the learning process. 



\subsubsection{Intransigence Measure ($I$)}
We define {\em intransigence} as the inability of a model to
learn new tasks. 
The effect of intransigence is more prominent in the
single-head setting especially in
the absence of previous data, as the model is expected to 
learn to differentiate the current task from the previous ones. 
Experimentally we show that storing
just a few representative samples (refer Sec.~\ref{sec:sampling}) from the
previous tasks improves intransigence significantly.
%
\noindent
Since we wish to quantify the {\em inability} to learn, we 
compare to the standard classification model which has access to all the
datasets at all times. We train a reference/target model with dataset \(
\bigcup_{l=1}^k\dataset_l \) and measure its accuracy on the held-out set of the
$k$-th task, denoted as $a_{k}^*$. We then define the intransigence for the $k$-th task as:
	\begin{equation}
		\label{eq:nlm_msr}
		I_k = a_{k}^* - a_{k,k}\ ,
	\end{equation}
where $a_{k,k}$ denotes the accuracy on the $k$-th task when trained up to task
$k$ in an incremental manner. Note, $I_k\in[-1,1]$,
and lower the $I_k$ the better the model. 
A reasonable reference/target model can be defined depending on the
feasibility to obtain it. In situations where it is highly expensive,
an approximation can be proposed.
\paragraph{Positive/Negative Forward Transfer ((P/N)FT):} 
Since intransigence is defined as the gap between the accuracy of an {\il}
algorithm and the reference model, negative intransigence ($I_k < 0$) implies
learning incrementally up to task $k$ positively influences model's
knowledge about it, \ie, positive forward transfer (PFT). Similarly, $I_k > 0$ implies
NFT. However, in~\cite{lopez2017gradient}, FT is quantified as the gain in
accuracy compared to the random guess (not a measure of intransigence) which is complementary to our approach.

\vspace{-0.4cm}
\section{Riemannian Walk for Incremental Learning}
\vspace{-0.3cm}
We first describe \textsc{ewc++}, an efficient version of the well known \textsc{ewc}~\cite{Kirkpatrick2016EWC}, and then RWalk which is a generalization of \textsc{ewc++} and \textsc{pi}~\cite{Zenke2017Continual}. Briefly, RWalk has three key components:
(1) a KL-divergence-based regularization over the conditional likelihood
$p_{\param}(\op|\ip)$ (\ewcp); (2) a parameter importance score based on the sensitivity
of the loss over the movement on the Riemannian manifold (similar to \textsc{pi}); and (3) strategies to obtain a few representative samples from 
the previous tasks. The first two components mitigate the
effects of catastrophic forgetting, whereas the third handles
intransigence.


\subsection{Avoiding Catastrophic Forgetting}
\label{sec:avoid_catast_forget}
\subsubsection{KL-divergence based Regularization \textsc{(ewc++)}}
\label{sec:klReg} We learn parameters for the current task such
that the new conditional likelihood is close (in terms of KL) to the one learned
until previous tasks. To achieve this, we regularize over the conditional
likelihood distributions $p_{\param}(\op|\ip)$ using the approximate
KL-divergence, Eq.~\eqref{eq:klapproxdiag1}, as the distance measure.
This regularization would preserve the inherent properties of the model about previous tasks as
the learning progresses.
Thus, given parameters $\theta^{k-1}$ trained sequentially from task $1$ to
$k-1$, and dataset $\dataset_k$ for the $k$-th task, the objective is:
\begin{equation}
    \label{eq:klonlyLoss}
    \argmin_{\theta} \tilde{L}^k (\theta) := L^k (\theta) +  \lambda\,
    \KL\left(p_{\param^{k-1}}(\op | \ip) \| p_{\param}(\op | \ip)\right)\ ,
\end{equation}
where, \( \lambda \) is a hyperparameter. Substituting
Eq.~\eqref{eq:klapproxdiag1}, the KL-divergence component can
be written as \( \KL\left(p_{\param^{k-1}} \| p_{\param}\right) \approx
\frac{1}{2} \sum_{i=1}^{\paramdim} F_{\param_i^{k-1}} (\theta_i - \theta_i^{k-1})^2\  \).
Note that, for two tasks, the above regularization is exactly the same as that
of {\ewc}~\cite{Kirkpatrick2016EWC}.
Here we presented it from the KL-divergence based perspective. 
Another way to look at it would be to consider Fisher\footnote{By Fisher we always mean the empirical Fisher information matrix.} for each parameter to be
its importance score. The intuitive explanation for this is as follows; since
Fisher captures the local curvature of the KL-divergence surface of the
likelihood distribution (as it is the component of the  second-order term of Taylor
approximation, refer Sec.~\ref{sec:kl}), higher Fisher implies higher curvature,
thus suggests to move less in that direction in order to preserve the likelihood.

In the case of multiple tasks, {\ewc} requires
storing Fisher for each task independently ($\bigoh(k\paramdim)$ parameters), and regularizing over all of them jointly. This 
is practically infeasible if there are many tasks and the network has millions of parameters. Moreover, 
to estimate the empirical Fisher, {\ewc} requires an additional pass at the end of training over
the dataset of each task (see Eq.~\eqref{eq:fisher}). To address these
two issues, we propose {\ewcp} that (1) maintains single diagonal Fisher matrix as the
training over tasks progresses, and (2) uses moving average for its efficient
update similar to~\cite{martens2015optimizing}. 
Given $F_{\param}^{t-1}$ at $t-1$, Fisher in {\ewcp} is updated as:
\begin{equation}\label{eq:moving_avg} 
F_{\param}^t = \alpha F_{\param}^t +
(1-\alpha)F_{\param}^{t-1}\ ,
\end{equation}
 where $F_{\param}^t$ is the Fisher matrix obtained
using the {\em current batch} and $\alpha \in [0,1]$ is a hyperparameter. Note,
$t$ represents the training iterations, thus, computing Fisher in this manner
contains information about previous tasks, and also eliminates the additional
forward-backward pass over the dataset. At the end of each task, we
simply store $F_{\param}^t$ as $F_{\param^{k-1}}$ and use it to regularize the
next task, thus storing only two sets of Fisher at any instant during 
training, irrespective of the number of tasks. Similar to {\ewcp}, an
efficient version of {\ewc}, referred as online-{\ewc}, is concurrently
developed in~\cite{schwarz2018progress}.


\begin{wrapfigure}{r}{0.5\textwidth}
\vspace{-7ex}
    \begin{center}
        \includegraphics[width=0.45\textwidth]{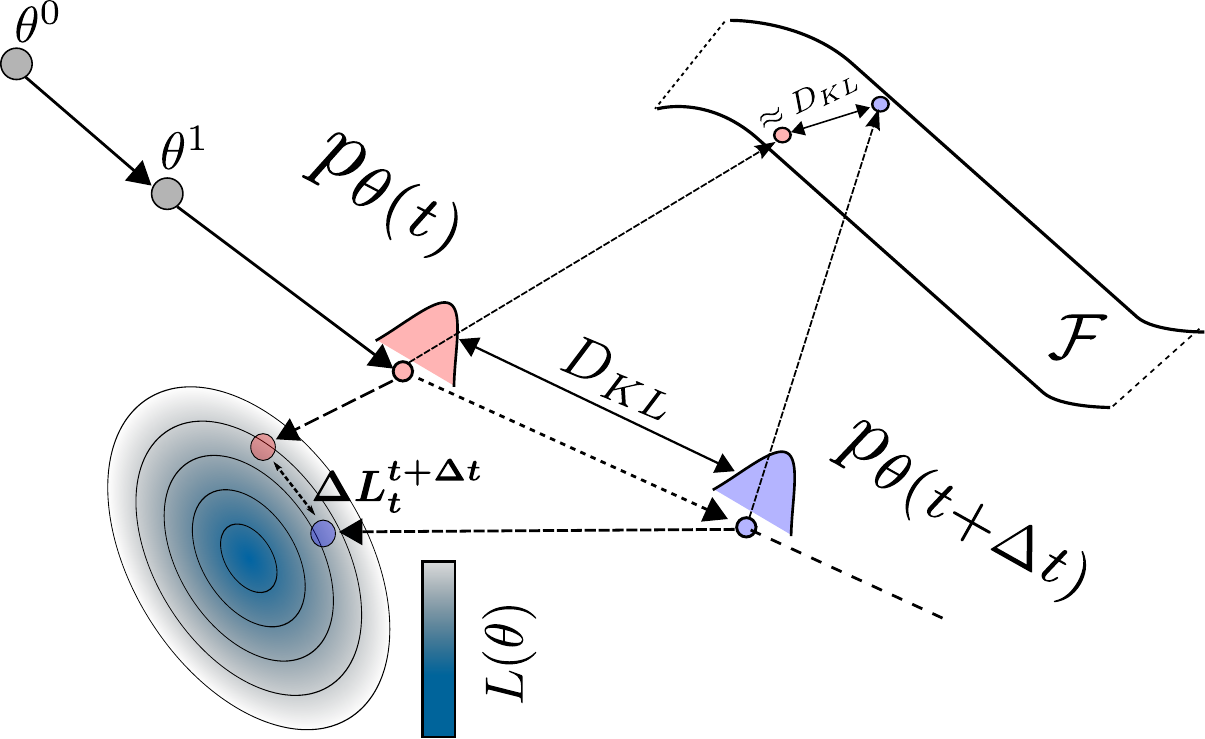}
        \vspace{-3ex}
    \end{center}
    \caption{\em Parameter importance accumulated over the optimization
trajectory.}
\label{fig:ourMethod}
\vspace{-5ex}
\end{wrapfigure}


In {\ewc}, Fisher is computed at a local minimum
of $\tilde{L}^k$ using the gradients of $L^k$, which is nearly zero whenever 
$\tilde{L}^k \approx L^k$ (\eg, smaller $\lambda$ or when $k=1$). 
This results in negligible regularization leading to catastrophic forgetting.
This issue is partially addressed in {\ewcp} using moving average. However, 
 to improve it further and to capture model's
 behaviour not just at the minimum
 but also during the entire training process, we augment each element
of the diagonal Fisher with a positive scalar as described below. This also
ensures that the augmented Fisher is always positive-definite.

\subsubsection{Optimization-path based Parameter Importance}
Since Fisher captures the intrinsic properties of the model and it only depends
on $L^k$,
it is blinded towards the influence of parameters over the optimization path on
the loss surface of $\tilde{L}^k$. We augment Fisher with a parameter importance score which is accumulated over the entire training trajectory of $\tilde{L}^k$ (similar to~\cite{Zenke2017Continual}).
This score is defined as the ratio of the change in the loss function
to the distance between the conditional likelihood distributions per
step in the parameter space.

More precisely, for a change of parameter from $\param_i(t)$
to $\param_i(t+1)$ (where $t$ is the time step or training iteration), we define
parameter importance as the ratio of the change in the loss to its influence in
$\KL(p_{\theta(t)} \| p_{\theta(t+1)})$. Intuitively, importance will be higher
if a small change in the distribution causes large
improvement over the loss. Formally, using the first-order Taylor approximation,
the change in loss \( \loss \) can be written as: 
\vspace{-0.2cm}
\begin{equation}
\label{eq:interm_Loss}
 \loss(\param(t + \Delta t)) \hspace{-2pt} - \loss(\param(t)) \hspace{-2pt} \approx \hspace{-2pt} -\sum_{i=1}^{\paramdim}
\hspace{-4pt} \sum_{t = t}^{t + \Delta t}  \hspace{-2pt} \frac{\partial \loss}{\partial \param_i} \left(\param_i(t+1) - \param_i(t)\right)  \hspace{-2pt} = \hspace{-3pt} -\sum_{i=1}^{\paramdim} \hspace{-2pt} \Delta \loss_{t}^{t + \Delta t}(\param_i),
\end{equation}
where $\frac{\partial \loss}{\partial \param_i}$ is
the gradient at $t$, and $\Delta
\loss_{t}^{t + \Delta t}(\param_i)$ represents the accumulated 
change in the loss caused by the change in the
parameter $\param_i$ from time step $t$ to
$t + \Delta t$. This change in parameter would cause a corresponding change in
the model distribution which can be computed using the approximate
KL-divergence (Eq.~\eqref{eq:klapproxdiag1}).
Thus, the importance of the parameter
$\param_i$ from training iteration $t_1$ to $t_2$ can be computed as
\(
    s_{t_1}^{t_2}(\param_i) = \sum_{t = t_1}^{t_2} \frac{\Delta \loss_{t}^{t +
    \Delta t}(\param_i)}{\frac{1}{2} F_{\theta_i}^t \Delta \theta_i(t)^2 + \epsilon} \ ,
\)
where $\Delta \theta_i(t) = \theta_i(t+\Delta t) - \theta_i(t)$ and $\epsilon >
0 $. The denominator is computed at every
discrete intervals of $\Delta t \geq 1$ and $F_{\theta_i}^t$ is computed efficiently at
every $t$-th step using moving average as described while explaining \textsc{ewc++}. 
The computation of this importance score is illustrated in
Fig.~\ref{fig:ourMethod}. Since we care about the positive influence of the parameters,
negative scores are set to zero. Note that, if the Euclidean distance 
is used instead, the score \( s_{t_1}^{t_2}(\param_i) \) would be
similar to that of {\pii}~\cite{Zenke2017Continual}.

\subsubsection{Final Objective Function (RWalk)} 
\label{sec:finalObj}
We now combine Fisher information matrix based importance and the optimization-path based
importance scores as follows:
\vspace{-0.25cm}
\begin{equation}
    \label{eq:auxLoss}
    \tilde{L}^k (\theta) = L^k (\theta) + \lambda\sum_{i=1}^{\paramdim}
    (F_{\param_i^{k-1}} + s_{t_0}^{t_{k-1}}(\param_i))(\theta_i - \theta_i^{k-1})^2\ .
\end{equation}
Here, $s_{t_0}^{t_{k-1}}(\param_i)$ is the score accumulated from the
first training iteration $t_0$ until the last training iteration $t_{k-1}$,
corresponding to task $k-1$. 
Since the scores are accumulated over time, the regularization
gets increasingly rigid. To alleviate this and enable continual learning,
after each task the scores are averaged: $s_{t_0}^{t_{k-1}}(\param_i) =
\frac{1}{2}\left(s_{t_0}^{t_{k-2}}(\param_i) +
s_{t_{k-2}}^{t_{k-1}}(\param_i)\right)$.
This continual averaging makes the tasks learned far in the past less influential than the tasks learned recently. Furthermore, while adding, it is important to make sure that the scales of both
\( F_{\param_i^{k-1}} \) and \( s_{t_0}^{t_{k-1}}(\param_i) \) are in the same order, so that the influence of
both the terms is retained. This can be ensured by individually normalizing them to be in the interval $[0,1]$. {\em This, together with score
averaging, have a positive side-effect of the regularization hyperparameter
$\lambda$ being less sensitive to the number of tasks.}
Whereas, {\ewc}~\cite{Kirkpatrick2016EWC} and {\pii}~\cite{Zenke2017Continual}
are highly sensitive to $\lambda$, making them relatively less reliable for
{\il}.
Note, during training, the space complexity for RWalk is $\bigoh(\paramdim)$,
independent of the number of tasks.
%

\subsection{Handling Intransigence}\label{sec:sampling}
Experimentally, we observed that training $k$-th task with $\dataset_k$ leads to
a poor test accuracy for the current task compared to previous tasks in the {\em
single-head} evaluation setting (refer Sec.~\ref{sec:heads}).
This happens because during training the model has access to $\dataset_k$ which contains labels 
only for the $k$-th task, $\op^k$. However, at test time the label space is over
all the tasks seen so far $\opspace^k = \cup_{j=1}^k \op^j$, which is much
larger than $\op^k$. This in turn increases {\em confusion} at test time
as the predictor function has no means to differentiate the samples of
the current task from the ones of previous tasks. An intuitive solution to
this problem is to store a small subset of representative samples from the previous
tasks and use it while training the current task~\cite{Rebuffi16icarl}. 
Below we discuss different strategies to obtain such a subset. 
Note that we store $m$ points from each task-specific dataset as the training
progresses, however, it is trivial to have a fixed total number of samples for
all the tasks similar to iCaRL~\cite{Rebuffi16icarl}.

\SKIP{
As a consequence to which the parameters do not update themselves prevents the network from forgetting on the classes seen previously but it does not allow it to discriminate current set of classes from the previous ones because at training time only the discriminative weights (parameters in the last layer of the neural network) of the current classes are being updated, and at test time the task is to assign a sample to one of the all the classes seen so far; $\opspace^k = \cup_{i=1}^k \op^i$ (section~\ref{sec:probSetup}). On the other hand, if one tries to update the discriminative weights of the previous classes while training for the current task, the model quickly suffers from the catastrophic forgetting despite the regularization loss as it does not see any samples from the previous classes.  

To learn a discriminator that can differentiate among different tasks, one possible strategy is to store a subset of representative samples from the previous data~\cite{lopez2017gradient,Rebuffi16icarl}. 
}

\subsubsection{Uniform Sampling} A na\"ive yet highly effective (shown experimentally)
approach is to sample uniformly at random from the previous datasets. 
\subsubsection{Plane Distance-based Sampling} In this case, we assume that samples
closer to the decision boundary are more representative than the ones far away.
For a given sample $\{\ip_i, y_i\}$, we compute the
pseudo-distance from the decision boundary $d(\ip_i) = \phi(\ip_i)^{\top} w^{y_i}$, where
$\phi(\cdot)$ is the feature mapping learned by the neural network and $w^{y_i}$ are
the last fully connected layer parameters for class $y_i$. Then, we sample points
based on $q(\ip_i) \propto \frac{1}{d(\ip_i)}$. Here, the intuition is, since the 
change in parameters is regularized, the feature space and the decision boundaries
do not vary much. Hence, the samples that lie close to the boundary would act as \textit{boundary defining samples}.
\subsubsection{Entropy-based Sampling} Given a sample, the entropy of the
output soft-max distribution measures the uncertainty of the sample which we
used to sample points. The higher the entropy the more likely is that the sample would be picked.
\subsubsection{Mean of Features (MoF)} iCaRL~\cite{Rebuffi16icarl} proposes a method
to find samples based on the feature space $\phi(\cdot)$. For each
class $y$, $m$ number of points are found whose mean in the feature space 
closely approximate the mean of the entire dataset for that class. However, this subset selection strategy is inefficient compared
to the above sampling methods.
In fact, the time complexity is $\bigoh(nfm)$ where $n$ is dataset size, $f$ is the feature
dimension and $m$ is the number of required samples. 
\section{Related Work}
One way to address catastrophic forgetting is by dynamically expanding
the network for each new task~\cite{lee2017lifelong,Rebuffi17,rusu2016progressive,terekhov2015knowledge}.
Though intuitive and simple, these approaches are not scalable as the size of the network
increases with the number of tasks. A better strategy would be to exploit the
over-parametrization of neural networks~\cite{hecht1988theory}. This entails regularizing 
either over the activations (output)~\cite{Rebuffi16icarl,li2016learning} or
over the network parameters~\cite{Kirkpatrick2016EWC,Zenke2017Continual}.
Even though activation-based approach allows more
flexibility in parameter updates, it is memory inefficient if the activations are in millions,
\eg, semantic segmentation. On the contrary,
methods that regularize over the parameters - weighting the parameters
based on their individual {\em importance} - are suitable for such tasks.
Our method falls under the latter category and we show that our method is a
generalization of {\ewcp} and {\pii}~\cite{Zenke2017Continual}, where {\ewcp} is
our efficient version of {\ewc}~\cite{Kirkpatrick2016EWC}, very similar to the
concurrently developed online-{\ewc}~\cite{schwarz2018progress}.
Similar in
spirit to regularization over the parameters, Lee~\etal~\cite{lee2017IMM} use moment matching to obtain network
weights as the combination of the weights of all the tasks, and 
Nguyen~\etal~\cite{nguyen2017variational} enforce the distribution over the
model parameters to be close via a Bayesian framework.
Different from the above approaches, Lopez-Paz~\etal~\cite{lopez2017gradient}
update gradients such that the losses of the previous tasks do not increase, while
Shin~\etal~\cite{shin2017continual} resort to a retraining strategy where the
samples of the previous tasks are generated using a learned generative model.



\section{Experiments}\label{sec:experiments}
%
\SKIP{
Even though we present results on both {\em multi-head} and {\em single-head}
evaluations (refer Sec.~\ref{sec:heads}), we advocate {\em single-head} over
{\em multi-head} because of its practicality, and experimentally
show that it is very hard to {\em preserve} and {\em update} knowledge for the
single-head setting while learning in an incremental manner.}

\SKIP{
\red{Note, {\em to consolidate our baselines for the single-head setting}, we make use of samples
from previous tasks for them as well and show that even a small amount of memory samples is sufficient to
considerably improve the baselines. Thus, making baselines much stronger than their original forms.}}
\subsubsection{Datasets}
We evaluate baselines and our proposed model - RWalk - on two datasets:
\begin{enumerate}
\item \textit{Incremental MNIST}: The standard MNIST dataset is split into five
disjoint subsets (tasks) of two consecutive
digits, \ie, $\cup_{k}\op^k = \{\{0,1\},\ldots, \{8,9\}\}$.
\item \textit{Incremental CIFAR}: To show that our approach scales to bigger
datasets, we use incremental CIFAR where CIFAR-100 dataset is split into ten
disjoint subsets such that $\cup_{k}\op^k = \{\{0-9\},\ldots, \{90-99\}\}$.
\end{enumerate}

\subsubsection{Architectures}
The architectures used are similar to~\cite{Zenke2017Continual}. For MNIST, we use an
MLP with two hidden layers each having $256$ units with ReLU
nonlinearites. For CIFAR-100, we use a CNN with
four convolutional layers followed by a single dense layer (see supplementary for more details). In all experiments, we use Adam optimizer~\cite{kingma2014adam} (learning rate = $1\times10^{-3}$, $\beta_1 =
0.9$, $\beta_2 = 0.999$) with a fixed batch size of $64$. 

\begin{table}[t]
\centering
\caption{\em Comparison with different baselines on MNIST and CIFAR in both
multi-head and single-head evaluation settings. Baselines where samples are
used are appended with '-S'. For MNIST and CIFAR, 10 (0.2\%) and 25(5\%) samples are used
from the previous tasks using mean of features (MoF) based sampling strategy (refer
Sec.~\ref{sec:sampling}).}
\label{tab:main_comparison}
\begin{tabular}{@{}c@{\hspace{8pt}}c@{\hspace{4pt}}c@{\hspace{4pt}}c@{\hspace{4pt}}c@{\hspace{4pt}}c@{\hspace{4pt}}c@{\hspace{4pt}}c@{\hspace{4pt}}c@{}}
\toprule
\multicolumn{1}{c}{\textbf{Methods}} &\multicolumn{4}{c}{\textbf{MNIST}} &\multicolumn{4}{c}{\textbf{CIFAR}} \\
\midrule
\multicolumn{9}{c}{Multi-head Evaluation}                                \\
\cmidrule{2-9}
                              & $\lambda$  & $A_5$(\%)        & $F_5$   & $I_5$ & $\lambda$       & $A_{10}$(\%)      & $F_{10}$ & $I_{10}$ \\

\cmidrule(r){2-5} \cmidrule(l){6-9}
Vanilla                       & 0          & 90.3  & 0.12    & $6.6\times10^{-4}$ & 0    & 44.4   & 0.36     & 0.02     \\
{\ewc} & 75000      & \textbf{99.3}  & 0.001  
& 0.01       & $3\times10^6$ & 72.8   & 0.001    & 0.07     \\
{\pii}  & 0.1        & \textbf{99.3}  & 0.002  
& 0.01       & 10            & 73.2   & 0        & 0.06    \\
\textbf{RWalk (Ours)}         & 1000       & \textbf{99.3}  & 0.003  
& 0.01       & 1000          & \textbf{74.2}  & 0.004    & 0.04    \\ \hline \multicolumn{9}{c}{Single-head Evaluation}                               \\
\cmidrule(r){2-5} \cmidrule(l){6-9}
\cmidrule{2-9} 
Vanilla                         & 0          & 38.0  & 0.62    & 0.29    & 0             & 10.2   & 0.36     & -0.06     \\
{\ewc}   & 75000      & 55.8  & 0.08    & 0.77    & $3\times10^6$ & 23.1   & 0.03     & 0.17     \\
{\pii}    & 0.1        & 57.6  & 0.11    & 0.8     & 10            & 22.8   & 0.04     & 0.2     \\
iCaRL-hb1 & -          & 36.6  & 0.68    & -0.01   & -             & 7.4    & 0.40     & 0.06     \\
iCaRL     & -          & 55.8  & 0.19    & 0.46    & -             & 9.5    & 0.11     & 0.35    \\
\cmidrule(r){2-5} \cmidrule(l){6-9}
Vanilla-S             & 0          & 73.7        & 0.30    & 0.03    & 0                 & 12.9   & 0.64     & -0.3     \\
{\ewc}-S                 & 75000      & 79.7        & 0.14    & 0.22    & $15\times10^5$    & 33.6   & 0.27     & -0.05    \\
{\pii}-S                  & 0.1        & 78.7        & 0.24    & 0.05    & 10                & 33.6   & 0.27     &  -0.03   \\
\textbf{RWalk (Ours)} & 1000       & \textbf{82.5}   & 0.15    & 0.14    & 500           & \textbf{34.0}   & 0.28     &  -0.06   \\
\bottomrule
\end{tabular}
\end{table}

\subsubsection{Baselines}
We compare RWalk against the following baselines:
\begin{itemize}
    \item Vanilla: Network trained without any regularization over past tasks.
    \item {\ewc}~\cite{Kirkpatrick2016EWC} and
    {\pii}~\cite{Zenke2017Continual}: 
    Both use parameter based regularization. 
    Note, we observed that {\ewcp} performed
    at least as good as {\ewc} and therefore, in all the experiments, by {\ewc} we mean
    the stronger baseline {\ewcp}.
    \item iCaRL~\cite{Rebuffi16icarl}: Uses regularization over the
    activations and a nearest-exempler-based classifier. Here, iCaRL-hb1
    refers to the {\em hybrid1} version, which uses the standard neural
    network classifier. Both the versions use previous samples.
\end{itemize}

Note, we use a few samples from the previous tasks to consolidate our baselines further in the single-head setting.
\SKIP{
\red{As mentioned earlier, for single-head setting, we create much stronger baselines to compare with by making use of samples from the previous tasks for them as well.}}

\subsection{Results}

\begin{figure}[t]
	\begin{subfigure}{0.40\linewidth}
		\includegraphics[scale=0.33]{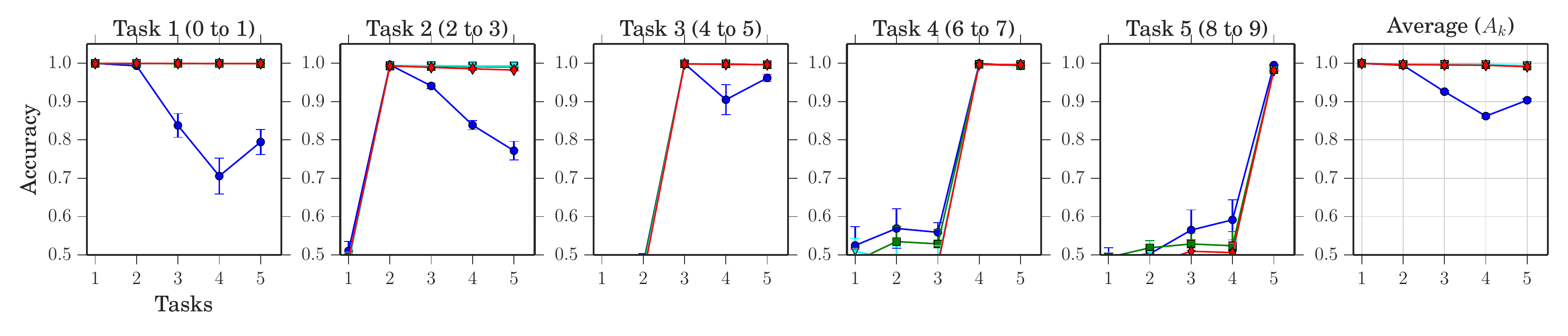}
	\end{subfigure}%

    \begin{subfigure}{0.40\linewidth}
		\includegraphics[scale=0.33]{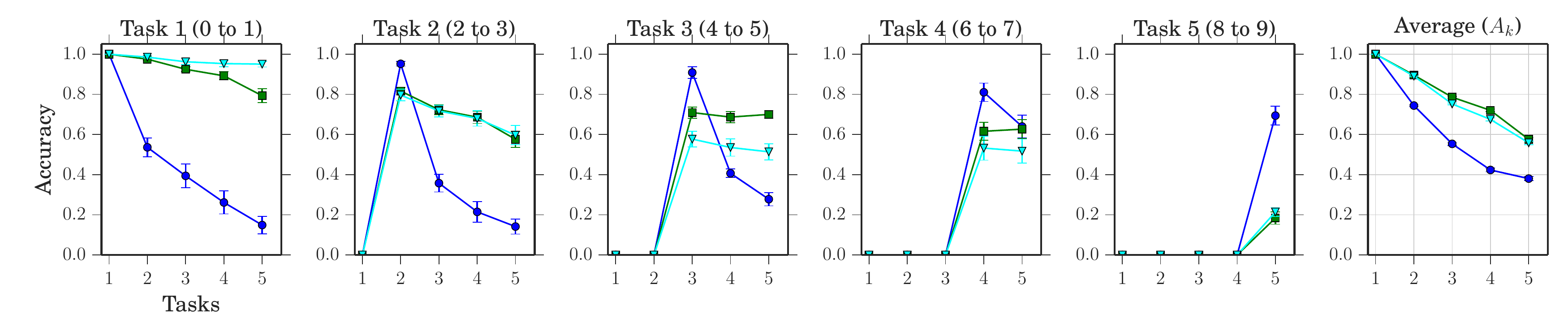}
	\end{subfigure}%
	
	\begin{subfigure}{0.40\linewidth}
		\includegraphics[scale=0.33]{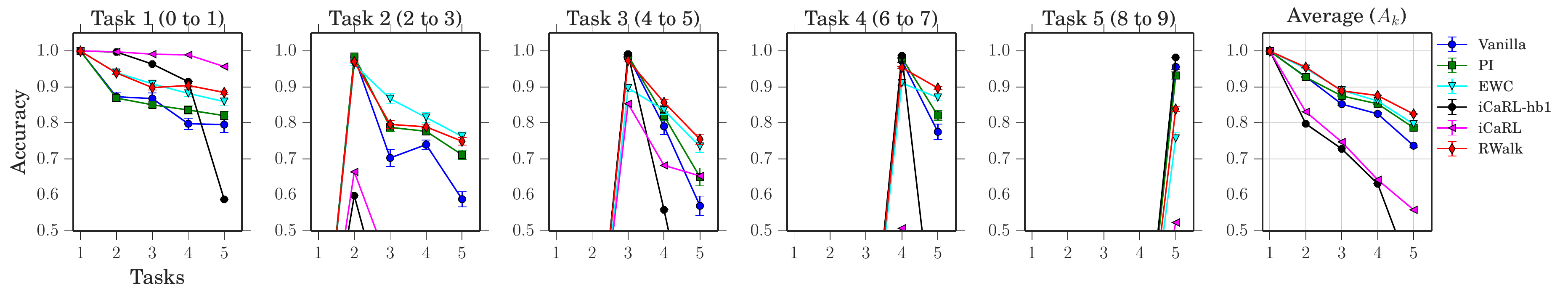}
	\end{subfigure}
\caption{\em Accuracy on incremental MNIST with multi-head evaluation
(\textbf{top}), and single-head evaluation without (\textbf{middle})
and with samples (\textbf{bottom}). First five columns show the variation in
performance for different tasks, \eg,
the first plot depicts the performance variation on Task 1 when trained
incrementally over five tasks. The last
column shows the accuracy ($A_k$, refer Sec.~\ref{sec:eval_measures}). 
Mean of features (MoF) sampling is used. 
}
\label{fig:mnist_acc_comp}
\end{figure}

We report the results in Tab.~\ref{tab:main_comparison} where RWalk outperforms all the baselines in terms of average accuracy and provides better trade-off between forgetting and intransigence. We now discuss the results in detail.

In the
multi-head evaluation setting~\cite{Zenke2017Continual,lopez2017gradient}, except Vanilla, all the methods provide state-of-the-art accuracy with {\em almost zero} forgetting and intransigence (top row of Fig.~\ref{fig:mnist_acc_comp}).
 {\em This gives an impression that
{\il} problem is solved.} However, as discussed in
Sec.~\ref{sec:heads}, this is an easier evaluation setting
and does not capture the essence of {\il}. 

However, in the single-head evaluation, {\em forgetting} and \textit{intransigence} increase substantially due to the 
the inability of the network to differentiate among tasks. 
Hence, the performance significantly drops for all the methods (refer Tab.~\ref{tab:main_comparison} and the middle row of Fig.~\ref{fig:mnist_acc_comp}).
For instance, on MNIST, forgetting and intransigence of Vanilla deteriorates from $0.12$ to $0.62$, and $6.6\times10^{-4}$ to $0.29$, respectively, causing the average accuracy to drop from $90.3$\% to $38.0$\%. Although, regularized methods, {\ewc} and {\pii}, designed to counter catastrophic forgetting, result in less degradation of forgetting, their accuracy is still significantly worse - compare $99.3$\% of {\pii} in multi-head against $57.6$\% in single-head.
In Tab.~\ref{tab:main_comparison}, a similar performance decrease is observed
on CIFAR-100 as well. 
Such a degradation in accuracy even with less forgetting shows that it is not only important to preserve knowledge (quantified by forgetting) but also to update knowledge (captured by intransigence) to achieve better performance. Task-level analysis for CIFAR dataset, similar to Fig.~\ref{fig:mnist_acc_comp}, is presented in the supplementary material.

We now show that even with a few representative samples intransigence can be mitigated.
For example, in the case of {\pii} on MNIST with only $10$ ($\approx 0.2$\%) samples for each previous class, the intransigence drops from $0.8$ to $0.05$ which results in improving the
average accuracy from $57.6$\% to $78.7$\%. Similar improvements can be seen for other methods as well.
On CIFAR-100, with only $5$\% representative samples, almost identical behaviour is observed.



In our CIFAR-100 experiments (CNN instead of ResNet32), we note that the performance of
iCaRL~\cite{Rebuffi16icarl} is significantly worse than what has been reported by the 
authors. 
We believe this is due to the dependence of iCaRL on a highly expressive feature space, as both the
regularization and the classifier depend on it. 
Perhaps, this reduced expressivity of the feature space due to the smaller network
resulted in the performance loss.



\begin{figure}[t]
\begin{center}
    \begin{subfigure}{0.48\linewidth}
        \begin{center}
        \includegraphics[scale=0.27]{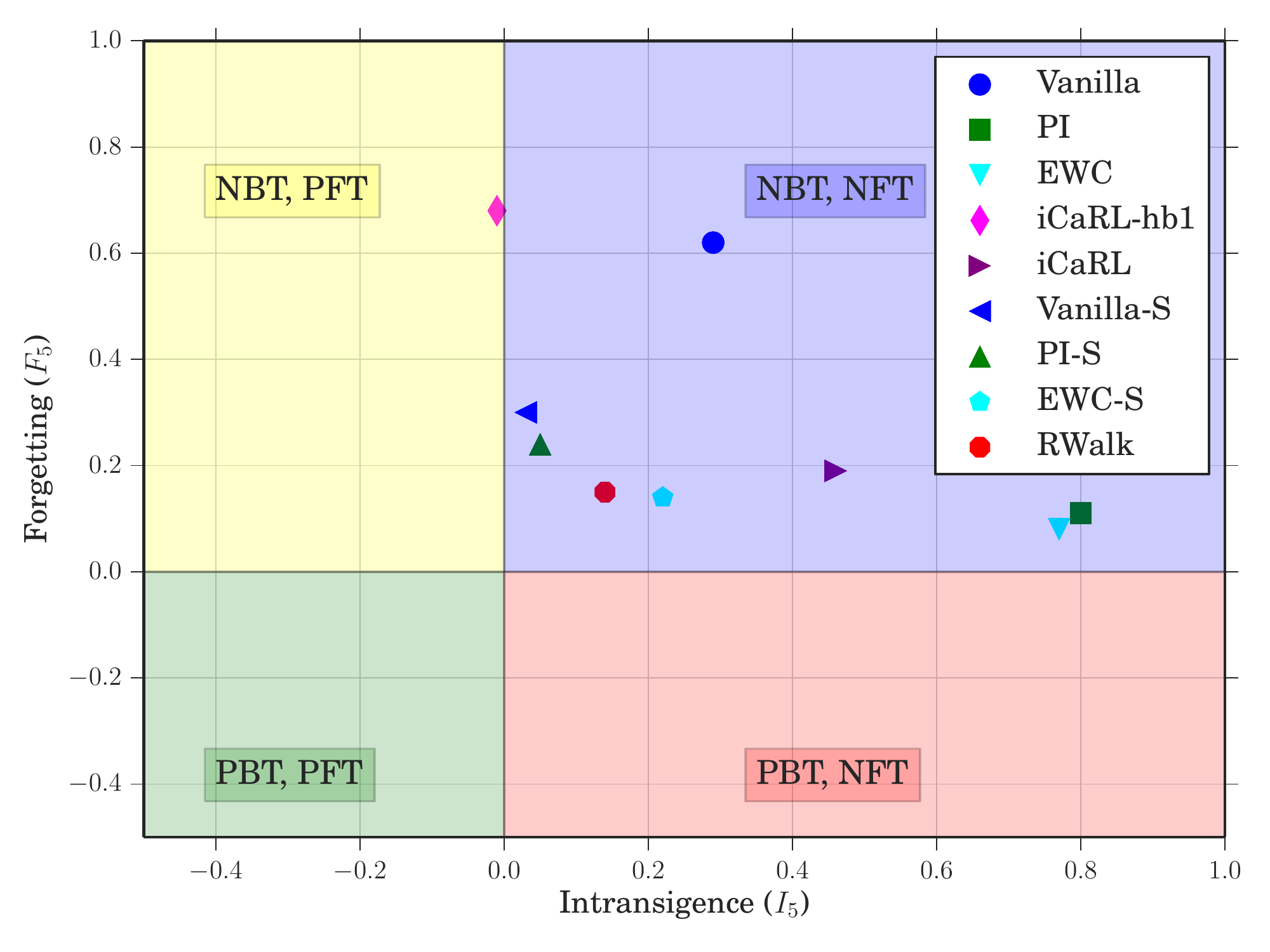}
        \caption{\small MNIST}
        \end{center}
    \end{subfigure}%
    \begin{subfigure}{0.48\linewidth}
        \begin{center}
        \includegraphics[scale=0.27]{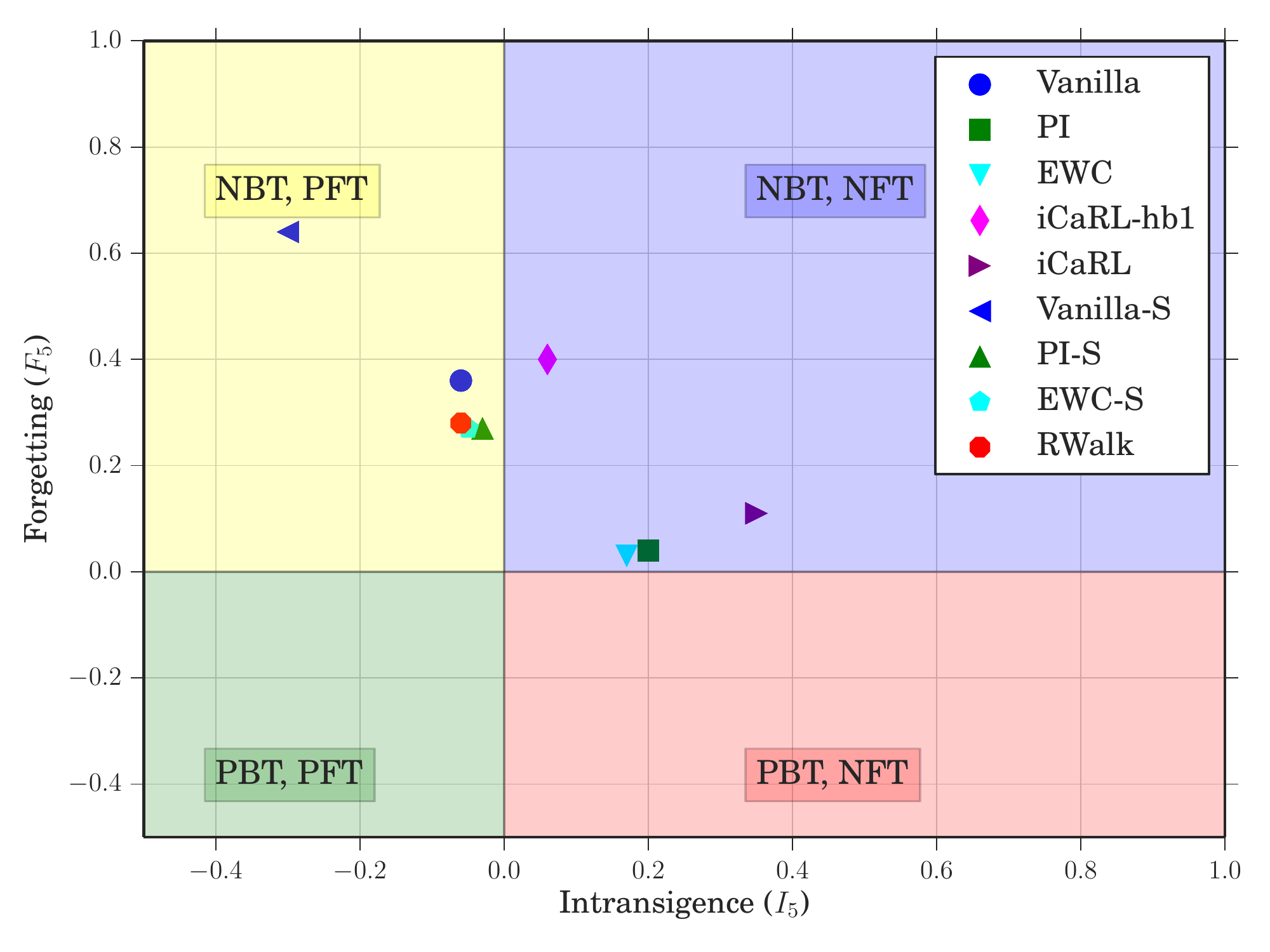}
        \caption{\small CIFAR}
        \end{center}
    \end{subfigure}
\end{center}
\vspace{-0.6cm}
\caption{\em Interplay between forgetting and intransigence.
}
\label{fig:scatter_fgt_intransigence}
\end{figure}

\begin{figure}[t]
\def \HEIGHT {25ex}
\def \SUBWIDTH {0.5\linewidth}
\begin{minipage}[c]{0.54\textwidth}
\begin{center}
    \begin{subfigure}{\SUBWIDTH}
        \begin{center}
        \includegraphics[height=\HEIGHT]{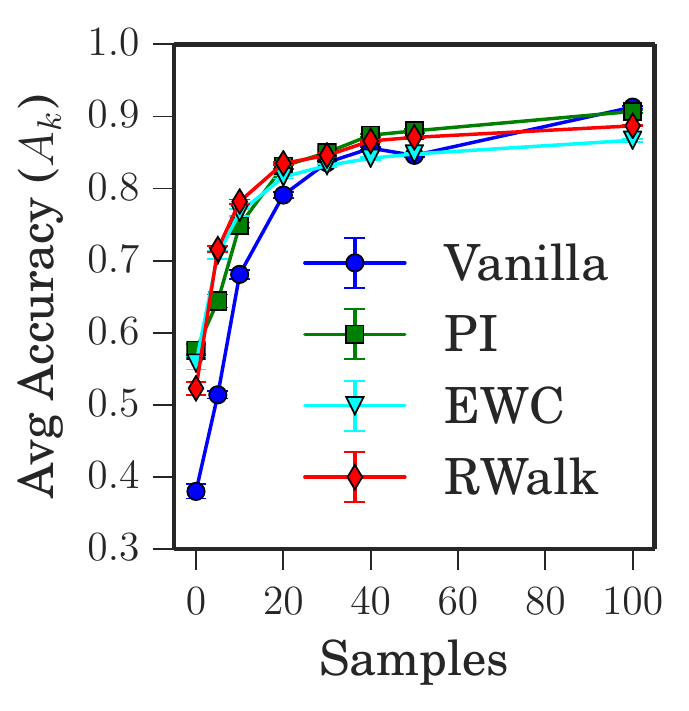}
        \vspace{-0.6cm}
        \caption{\small MNIST}
        \end{center}
    \end{subfigure}%
    \begin{subfigure}{\SUBWIDTH}
        \begin{center}
        \includegraphics[height=\HEIGHT]{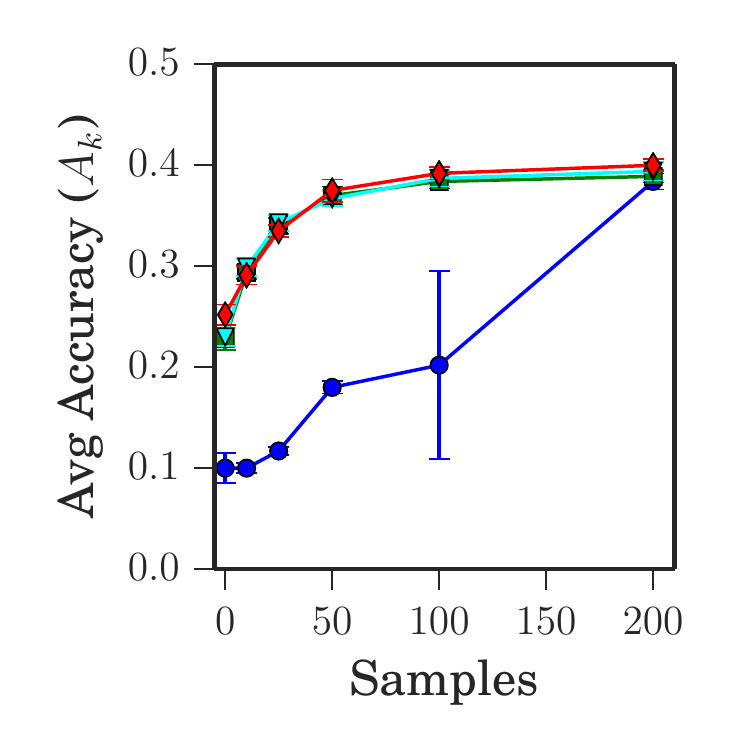}
        \vspace{-0.6cm}
        \caption{\small CIFAR}
        \end{center}
    \end{subfigure}
\end{center}
\end{minipage}\hfill
  \begin{minipage}[c]{0.425\textwidth}
\caption{\em Comparison by increasing the number of samples. 
On MNIST and CIFAR each class has around 5000 and 500 samples, respectively. 
With increasing number of samples, the performance of Vanilla improved, but in the range where 
Vanilla is poor, RWalk consistently performs the best. Uniform sampling is
used. 
}
\label{fig:increasing_sampling_comp}
 \end{minipage}
\end{figure}

\begin{figure}[t]
\def \HEIGHT {20ex} 
\def \SUBWIDTH {0.25\linewidth} 
\begin{center}
    \begin{subfigure}[t]{\SUBWIDTH}
        \begin{center}
        \includegraphics[height=\HEIGHT]{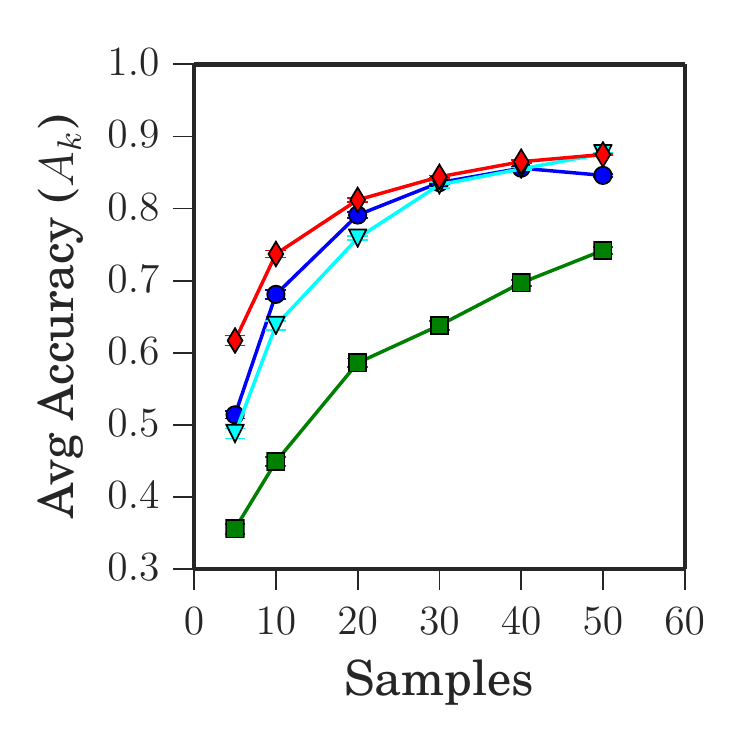}
        \end{center}
    \end{subfigure}%
    \begin{subfigure}[t]{\SUBWIDTH}
        \begin{center}
        \includegraphics[height=\HEIGHT]{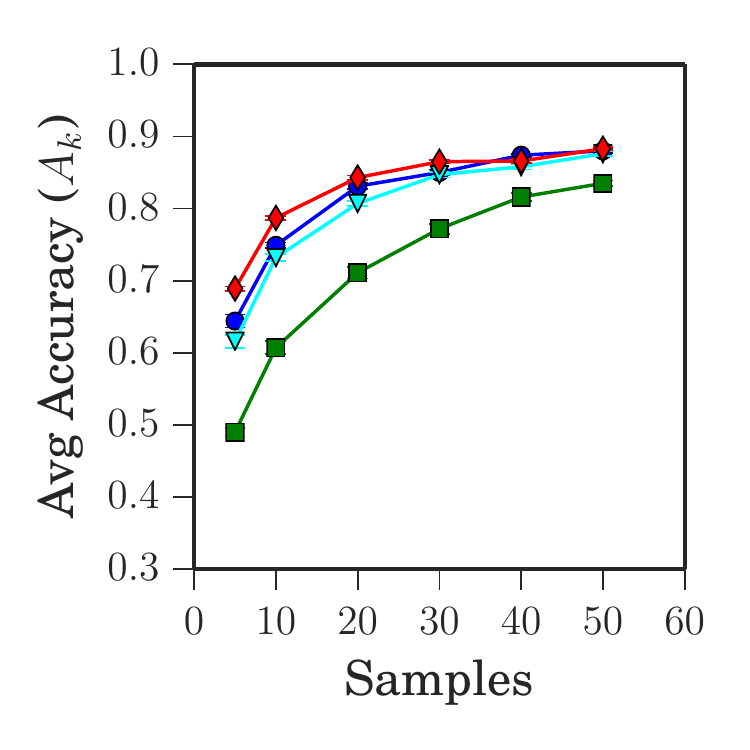}
        \end{center}
    \end{subfigure}%
    \begin{subfigure}[t]{\SUBWIDTH}
        \begin{center}
        \includegraphics[height=\HEIGHT]{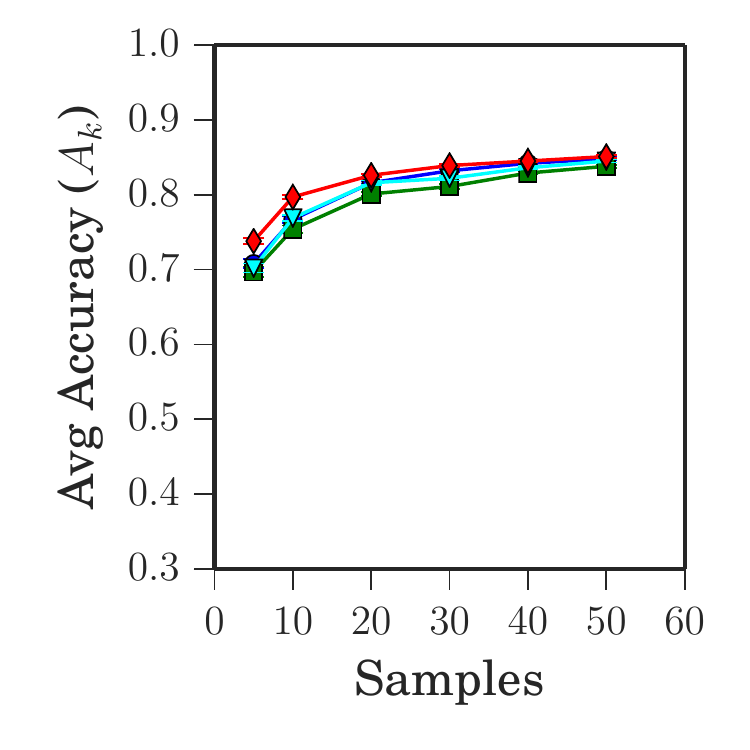}
        \end{center}
    \end{subfigure}%
    \begin{subfigure}[t]{\SUBWIDTH}
        \begin{center}
        \includegraphics[height=\HEIGHT]{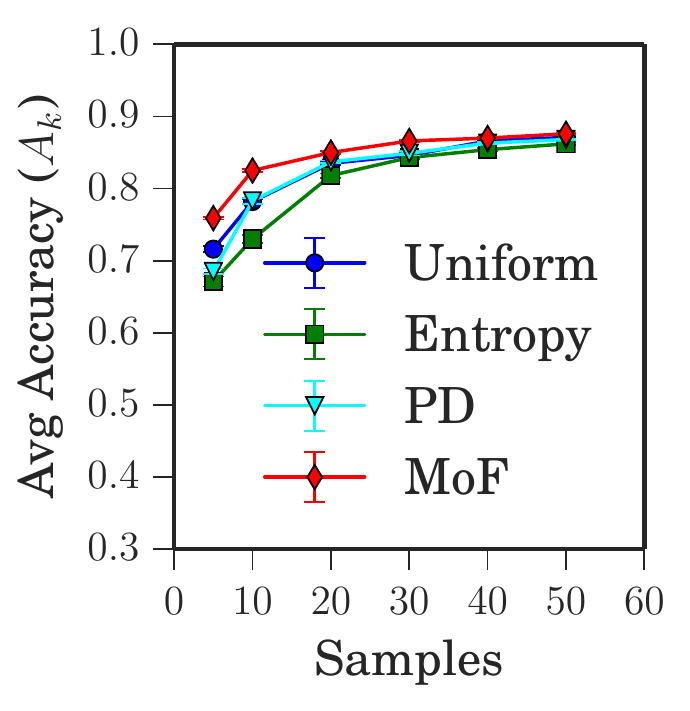}
        \end{center}
    \end{subfigure}
    
     \begin{subfigure}[t]{\SUBWIDTH}
        \begin{center}
        \includegraphics[height=\HEIGHT]{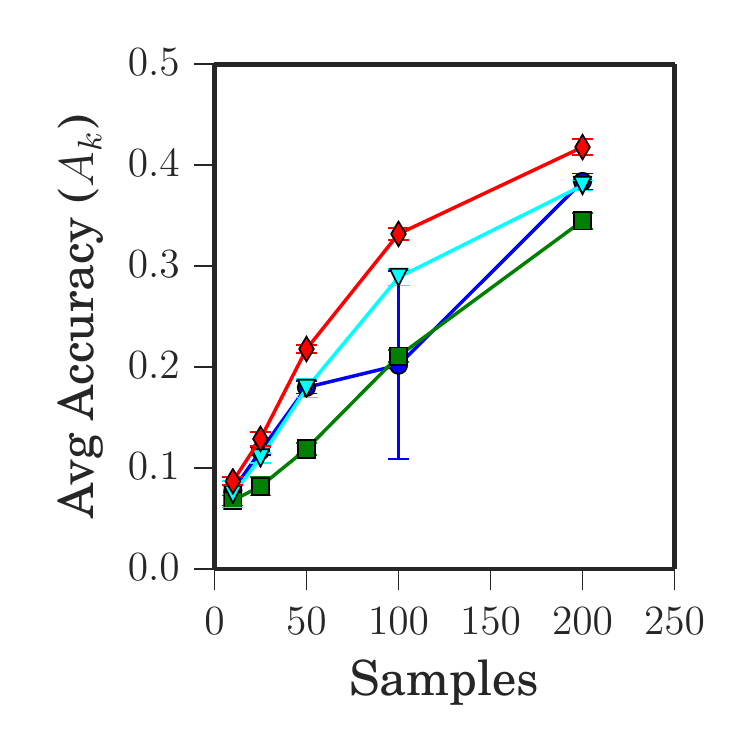}
        \caption{\small Vanilla}
        \end{center}
    \end{subfigure}%
    \begin{subfigure}[t]{\SUBWIDTH}
        \begin{center}
        \includegraphics[height=\HEIGHT]{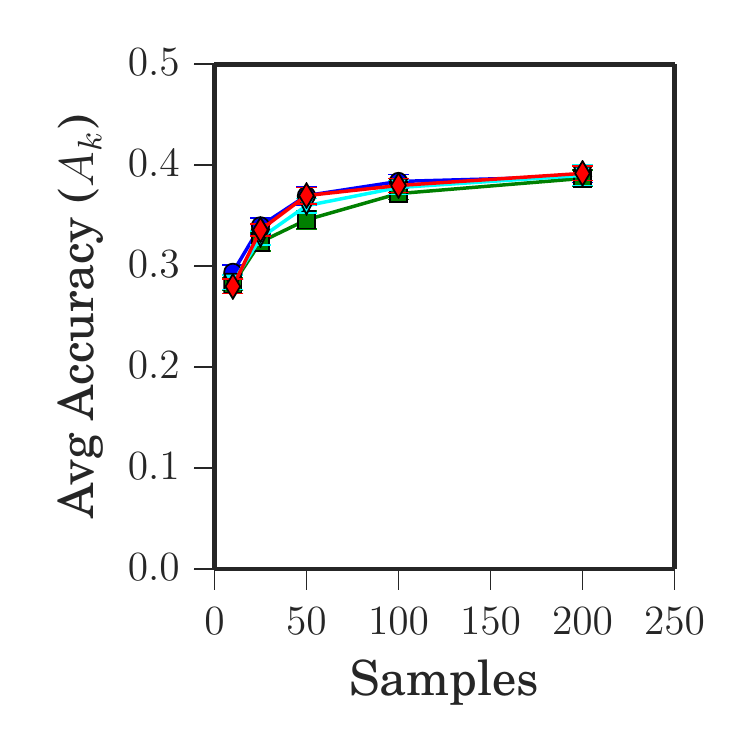}
        \caption{\small {\pii}}
        \end{center}
    \end{subfigure}%
    \begin{subfigure}[t]{\SUBWIDTH}
        \begin{center}
        \includegraphics[height=\HEIGHT]{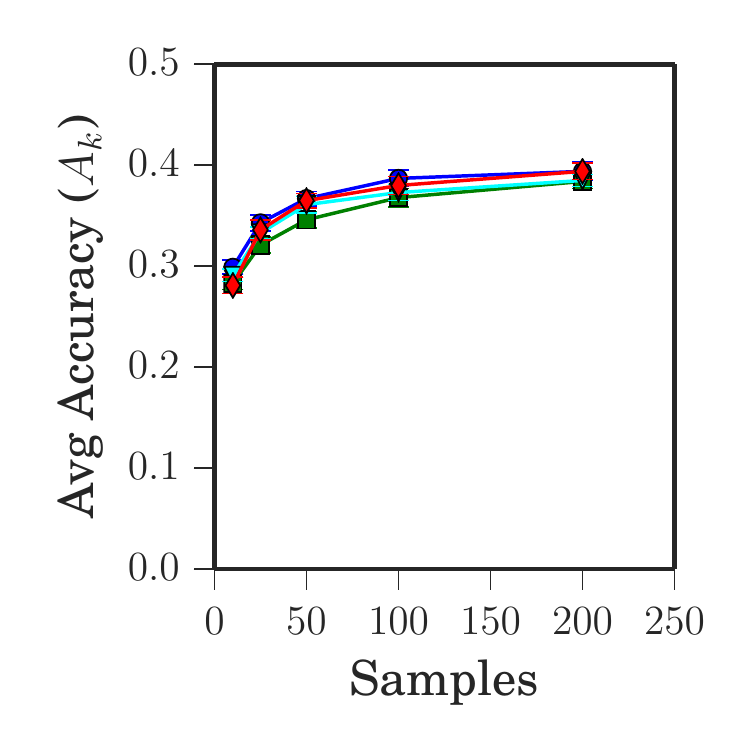}
        \caption{\small {\ewc}}
        \end{center}
    \end{subfigure}%
    \begin{subfigure}[t]{\SUBWIDTH}
        \begin{center}
        \includegraphics[height=\HEIGHT]{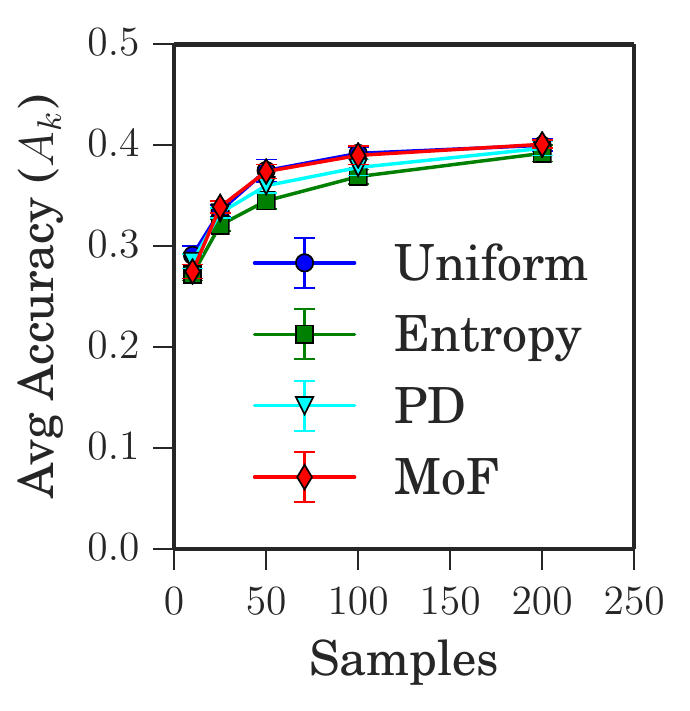}
        \caption{\small RWalk}
        \end{center}
    \end{subfigure}   
\end{center}
\caption{\em Comparison of different sampling strategies discussed in
Sec.~\ref{sec:sampling} on MNIST (\textbf{top}) and CIFAR-100
(\textbf{bottom}). Mean of features (MoF) outperforms others.
}
\label{fig:different_sampling_comp}
\end{figure}

\subsubsection{Interplay of Forgetting and Intransigence}
In Fig.~\ref{fig:scatter_fgt_intransigence} we study the interplay of
forgetting and intransigence in the single-head setting. Ideally we would like a model to be in the
quadrant marked as {\em PBT, PFT} (\ie, positive backward transfer and positive forward transfer). 
On MNIST, since all the methods, except iCaRL-hb1, lie
on the top-right quadrant, hence for models with comparable accuracy, a model which has the
smallest distance from $(0,0)$ would be better. 
As evident, RWalk is closest to $(0,0)$, providing a better trade-off between forgetting and intransigence compared to all the other methods. 
On CIFAR-100, the models lie on both the top quadrants and with the introduction
of samples, all the regularized methods show positive forward transfer. Since the models lie on different quadrants, their comparison of forgetting and
intransigence becomes application specific. In some cases, we might prefer a model that 
performs well on new tasks (better intransigence), irrespective of its performance on the old ones (can compromise forgetting), and vice versa. 
Note that, RWalk maintains comparable performance to other baselines while yielding higher
average accuracy on CIFAR-100.

\subsubsection{Effect of Increasing the Number of Samples}
As expected, for smaller number of samples, regularized methods perform far superior
compared to Vanilla (refer Fig.~\ref{fig:increasing_sampling_comp}). However, once the number of samples are sufficiently
large, Vanilla starts to perform better or equivalent to the regularized models.
The reason is simple because now the Vanilla has access to enough samples of the
previous tasks to relearn them at each step, thereby obviating the need of
regularized models. However, in an {\il} problem, a fixed
small-sized memory budget is usually assumed. Therefore, one cannot afford to store large
number of samples from previous tasks. Additionally, for a simpler dataset like MNIST, Vanilla quickly
catches up to the regularized models with small number of samples ($20$, 
$0.4$\% of total samples) but on a more challenging dataset like CIFAR it takes considerable
amount of samples ($200$, $40$\% of total samples) of previous tasks for Vanilla to match the
performance of the regularized models. 

\subsubsection{Comparison of Different Sampling Strategies}
In Fig.~\ref{fig:different_sampling_comp} we compare different subset
selection strategies discussed in Sec.~\ref{sec:sampling}. It can be observed that
for all the methods Mean-of-Features (MoF) subset selection procedure,
introduced in iCaRL~\cite{Rebuffi16icarl}, performs the best. Surprisingly, \textit{uniform} sampling, despite being simple, is as
good as more complex MoF, Plane Distance (PD) and entropy-based sampling
strategies.
Furthermore, the regularized methods remain insensitive to different
sampling strategies, whereas in Vanilla, performance varies a lot against
different strategies. We believe this is due to the unconstrained change in the
last layer weights of the previous tasks. 

\section{Discussion}
In this work, we analyzed the challenges in the incremental learning problem, 
namely, catastrophic forgetting and intransigence, and
introduced metrics to quantify them. Such metrics 
reflect the interplay between {\em forgetting} and
{\em intransigence},
which we believe will encourage future research for exploiting model
capacity, such as, sparsity enforcing regularization, and exploration-based
methods for incremental learning.
In addition, we have presented an efficient version of {\ewc} referred to as {\ewcp}, 
and a generalization of 
{\ewcp} and {\pii} with a KL-divergence-based
perspective. Experimentally, we observed that these parameter
regularization methods suffer from high intransigence in the practical 
{\em single-head} setting and showed that this can be alleviated with a small
subset of representative samples. Since these methods
are memory efficient compared to knowledge distillation-based algorithms such
as iCaRL, future research in this direction would
enable the possibility of incremental learning on segmentation tasks.

\SKIP{
In this work, we have critically analyzed the challenges in an incremental
learning problem, namely, catastrophic forgetting and intransigence. To tackle
catastrophic forgetting, we have presented an incremental learning algorithm
that regularizes the KL-divergence between the conditional probability
distributions defined by the parameters of the neural network. In addition to
that, to enhance flexibility in learning, a parameter importance, based on
sensitivity of the loss with respect to the change in KL, is introduced.
In addition to that, not-learning is handled using sampling 
strategies based on feature space and decision boundaries. 
Our complete algorithm is memory efficient compared to knowledge distillation
based algorithms such as iCaRL~\cite{Rebuffi16icarl}, which would enable the
possibility of incremental learning on segmentation tasks. 
Furthermore, we have 
introduced metrics to measure forgetting and intransigence, which we believe,
would be useful to better evaluate incremental learning algorithms.
}

\section*{Acknowledgements}
This work was supported by The Rhodes Trust, EPSRC, ERC grant ERC-2012-AdG 321162-HELIOS, EPSRC grant Seebibyte EP/M013774/1 and EPSRC/MURI grant EP/N019474/1. Thanks to N. Siddharth for suggesting the term `intransigence'.

\clearpage

\clearpage
\section*{Supplementary Material}
\appendix

For the sake of completeness, we first give more details on the KL-divergence approximation using Fisher information matrix 
(Sec.~\red{2.3}). In particular, we give the proof of KL approximation, \( \KL(p_{\param} \| p_{\param + \Delta \param}) \approx \frac{1}{2} \Delta \param^{\top} F_{\param} \Delta \param \), discuss the difference between the true Fisher and the empirical Fisher\footnote{By Fisher, we always mean the empirical Fisher.}, and explain why the Fisher goes to zero at a minimum. Later, in Sec.~\ref{sec:resnet_exper}, we provide a comparison with GEM~\cite{lopez2017gradient} and show that RWalk significantly outperforms it. Note that, comparison with GEM is not available in the main paper. 
Additionally, we discuss the sensitivity of different models to the regularization hyperparameter ($\lambda$) in Sec.~\ref{sec:lambda_exper}. Finally, we conclude in Sec.~\ref{sec:supp_network_arch} with the details of the architecture and task-based analysis of the network used for CIFAR-100 dataset.
 {\em We note that with additional experiments and further analysis in this supplementary the conclusions of the main paper hold.}

\section{Approximate KL divergence using Fisher Information Matrix}\label{sec:supp_kl}
\subsection{Proof of Approximate KL divergence}
\begin{lemma}
Assuming $\Delta \param \to 0$, the second-order Taylor approximation of
KL-divergence can be
written~\cite{Amari98NaturalGradient,Pascanu14NaturalGradient} as:
\begin{equation}
\label{eq:KLapprox}
    \KL(p_{\param} \| p_{\param + \Delta \param}) \approx
     \frac{1}{2} \Delta \param^{\top} {F}_{\param} \Delta \param\ ,
\end{equation}
where $F_{\param}$ is the empirical Fisher at $\param$.
\end{lemma}

\begin{proof}
The KL divergence is defined as:
\begin{equation}\label{eq:supp_kl}
\KL(p_{\param}(\mathbf{z}) \| p_{\param + \Delta \param}(\mathbf{z})) = \E_{\mathbf{z}}\left[\log p_{\param}(\mathbf{z}) - \log p_{\param + \Delta \param}(\mathbf{z})\right]\ .
\end{equation}
Note that we use the shorthands $p_{\param}(\mathbf{z}) = p_{\param}(\op | \ip)$ and $\E_{\mathbf{z}}[\cdot] = \E_{\ip \sim \dataset, \op \sim p_{\param}(\op | \ip)}[\cdot]$. We denote partial derivatives as column vectors.
Let us first write the second order Taylor series expansion of $\log p_{\param + \Delta \param}(\mathbf{z})$ at $\param$:
\begin{equation}
\label{eq:taylor_approx}
 \log p_{\param + \Delta \param} \approx \log p_{\param} + \Delta \param ^{\top}\frac{\partial \log p_{\param}}{\partial \param} + \frac{1}{2}\Delta \param^{\top} \frac{\partial^2 \log p_{\param}}{\partial \param^2} \Delta \param\ .
\end{equation}
Now, by substituting this in Eq.~\eqref{eq:supp_kl}, the KL divergence can be approximated as:
\begin{subequations}
\begin{align}
	\label{subeq:exp_zero}
&\KL(p_{\param} \| p_{\param + \Delta \param}) \approx \E_{\mathbf{z}}[\log p_{\param}] - \E_{\mathbf{z}}[\log p_{\param}] \\ \nonumber
	&\quad -  \Delta \param ^{\top} \E_{\mathbf{z}}\left[ \frac{\partial \log p_{\param}}{\partial \param}\right] - \frac{1}{2} \Delta \param^{\top} \E_{\mathbf{z}}\left[\frac{\partial^2 \log p_{\param}}{\partial \param^2}\right] \Delta \param\ ,\\\nonumber
	& = \frac{1}{2} \Delta \param^{\top} \E_{\mathbf{z}}\left[-\frac{\partial^2 \log p_{\param}}{\partial \param^2}\right] \Delta \param \quad \mbox{see Eq.~\eqref{eq:exp_zero}}\ ,\\
	\label{subeq:hessian_fisher}
	& = \frac{1}{2} \Delta \param^{\top} \variant{H} \Delta \param \quad \mbox{see Eq.~\eqref{subeq:hessian_fisher_b}}\ .
\end{align}
\end{subequations}
In Eq.~\eqref{subeq:exp_zero}, since the expectation is taken such that, $\ip \sim \dataset, \op \sim p_{\param}(\op | \ip)$, the first order partial derivatives cancel out, \ie,
\begin{align}\label{eq:exp_zero}
\E_{\mathbf{z}}\left[ \frac{\partial \log p_{\param}(\mathbf{z})}{\partial \param}\right] &= \E_{\ip \sim \dataset}\left[\sum_{\mathbf{\op}} p_{\param}(\mathbf{\op | \ip}) \frac{\partial \log p_{\param}(\op | \ip)}{\partial \param}\right]\ ,\\\nonumber
&= \E_{\ip \sim \dataset}\left[ \sum_{\mathbf{\op}} p_{\param}(\op | \ip) \frac{1}{p_{\param}(\op | \ip)} \frac{\partial p_{\param}(\op | \ip)}{\partial \param}\right] \ ,\\\nonumber
	& = \E_{\ip \sim \dataset}\left[ \frac{\partial}{\partial \param} \sum_{\mathbf{\op}} p_{\param}(\op | \ip)\right] \ ,\\\nonumber
	& = \E_{\ip \sim \dataset}[0] = 0 \ .
\end{align}
Note that this holds for the continuous case as well where, assuming sufficient smoothness and the fact that limits of integration are constants ($0$ to $1$), the Leibniz's rule would allow us to interchange the differentiation and integration operators.

Additionally, in Eq.~\eqref{subeq:hessian_fisher}, the expected value of negative of the Hessian can be shown to be equal to the true Fisher matrix ($\tilde{F}$) by using Information Matrix Equality. 
\begin{subequations}
\begin{align}
\label{subeq:hessian_fisher_a}
& \E_{\mathbf{z}}\left[-\frac{\partial^2 \log p_{\param}(\mathbf{z})}{\partial \param^2}\right] = - \E_{\mathbf{z}}\left[\frac{1}{p_{\param}(\mathbf{z})}\frac{\partial^2 p_{\param}(\mathbf{z})}{\partial \param^2}\right] \,\\\nonumber
&\quad + \E_{\mathbf{z}}\left[\left( \frac{\partial \log p_{\param}(\mathbf{z}) }{\partial \param} \right) \left( \frac{\partial \log p_{\param}(\mathbf{z}) }{\partial \param} \right)^{\top}\right] \ ,\\\
\label{subeq:hessian_fisher_b}
&= - \E_{\mathbf{z}}\left[\frac{1}{p_{\param}(\mathbf{z})}\frac{\partial^2 p_{\param}(\mathbf{z})}{\partial \param^2}\right] + \tilde{F}_{\param} \ .
\end{align}
\end{subequations}

\begin{itemize}
	\item By the definition of KL-divergence, the expectation in the above equation is taken such that, $\ip \sim \dataset, \op \sim p_{\param}(\op | \ip)$. This cancels out the first term by following a similar argument as in Eq.~\eqref{eq:exp_zero}. Hence, in this case, the expected value of negative of the Hessian equals true Fisher matrix ($\tilde{F}$).
	\item However, if in Eq.~\eqref{subeq:hessian_fisher_b}, the expectation is taken such that, $(\ip, \op) \sim \dataset$, the first term does not go to zero, and $\tilde{F}_{\param}$ becomes the \emph{empirical Fisher matrix} ($F_{\param}$).
	
	\item Additionally, at the optimum, since the model distribution approaches the true data distribution, hence even sampling from dataset \ie, $(\ip, \op) \sim \dataset$ will make the first term to approach zero, and $\variant{H} \approx F_{\param}$.
	 
\end{itemize}
With the approximation that $\variant{H} \approx \tilde{F}_{\param} \approx F_{\param}$, the proof is complete.

\end{proof}

We will argue in Section~\ref{sec:true_fisher} that the true Fisher matrix is expensive to compute as it requires multiple backward passes. Therefore, as mostly used in literature~\cite{Amari98NaturalGradient,Pascanu14NaturalGradient}, we also employ empirical Fisher approximation to obtain the KL-divergence. 
\subsection{Empirical vs True Fisher}
\paragraph{Loss gradient}
\label{sec:fisher_at_minimum}
Let $q$ be any reference distribution and $p$ (parametrized by $\param$) be the model distribution obtained after applying softmax on the class scores ($s$). The cross-entropy loss between $q$ and $p$ can be written as: $\ell(\param) = -\sum_j q_j \log p_j$. The gradients of the loss with respect to the class scores are:
\begin{equation}
\label{eq:grad_entropy}
	\frac{\partial \ell(\param)}{\partial s_j} = p_j - q_j\ .
\end{equation} 

\noindent 
By chain rule, the loss gradients w.r.t. the model parameters are \( \frac{\partial \ell(\param)}{\partial \param} = \frac{\partial \ell(\param)}{\partial \mathbf{s}}\frac{\partial \mathbf{s}}{\partial \param} \).

\subsubsection{Empirical Fisher}
\label{sec:emp_fisher}
In case of an empirical Fisher, the expectation is taken such that $(\ip, \op) \sim \dataset$. Since every input $\ip$ has only one ground truth label, this makes $q$ a Dirac delta distribution . Then, Eq.~\eqref{eq:grad_entropy} becomes:
\begin{equation*}
	\frac{\partial \ell(\param)}{\partial s_j} =
	\begin{cases}
      p_j - 1, & \text{if `$j$' is the ground truth label}\ , \\
      p_j, & \text{otherwise}\ .
    \end{cases}
\end{equation*}

Since {\em at any optimum} the loss-gradient approaches to zero, thus, Fisher being the expected loss-gradient covariance matrix would also approach to a zero matrix.

 
\subsubsection{True Fisher}
\label{sec:true_fisher}
In case of true Fisher, given $\ip$, the expectation is taken such that $\op$ is sampled from the model distributions $p_{\param}(\op | \ip )$. If the final layer of a neural network is a soft-max layer and the network is
trained using cross entropy loss, then the output may be interpreted as a
probability distribution over the categorical variables.
Thus, at a given \(\param\),
the conditional likelihood distribution learned by a neural network is  actually
a conditional multinoulli distribution defined as $p_{\param}(\op | \ip ) =
\prod_{j=1}^K p_{\param,j}^{[y=j]}$, where $p_{\param,j}$ is the soft-max
probability of the $j$-th class, $K$ are the total number of classes, $\op$ is
the one-hot encoding of length $K$, and $[\cdot]$ is Iverson bracket.
At a good optimum, the model distribution $p_{\param}(\op | \ip )$ becomes peaky around the ground truth label, implying $p_{\param,t} \gg p_{\param,j}, \forall j \neq t$ where $t$ is the ground-truth label. Thus, given input $\ip$, the model distribution $p_{\param}(\op | \ip )$ approach the ground-truth output distribution. This makes the true and empirical Fisher behave in a very similar manner.
Note, in order to compute the expectation over the model distribution, the true Fisher requires multiple backward passes making it prohibitively expensive to compute. The standard practice is to resort to the empirical Fisher approximation in this situation~\cite{Amari98NaturalGradient,Pascanu14NaturalGradient}.

\section{Additional Experiments and Analysis}
\subsection{Comparison with GEM~\cite{lopez2017gradient} on ResNets}
\label{sec:resnet_exper}
 
\begin{table}[h]
\centering
\caption{\em Following GEM, all the results on ResNets are in the multi-head evaluation setting. Note that, the total number of samples are from all the tasks combined.}
\label{tab:resnet_comp}
\begin{tabular}{c@{\hspace{10pt}}c@{\hspace{10pt}}c}
\toprule
\textbf{Methods}             & \textbf{Total Number of Samples} & \boldsymbol{$A_k$} (\%) \\ 
iCaRL  & 5120                             & 50.8            \\ 
GEM & 5120                             & 65.4             \\ 

{\textbf{RWalk (Ours)}}      & 5000                             & \textbf{70.1}      \\
\bottomrule
\end{tabular}
\end{table}

In this section we show experiments with ResNet18~\cite{he2016deep} on CIFAR-100 dataset. In Tab.~\ref{tab:resnet_comp} we report the results where we compare our method with iCaRL~\cite{Rebuffi16icarl} and Gradient Episodic Memory (GEM)~\cite{lopez2017gradient}. Both of these methods use ResNet18 as an underlying architecture. Following GEM-setup, we split the CIFAR-100 dataset in $20$ tasks where each task consists of $5$ consecutive classes, such that $\cup_{k=1}^{20}\op^k = \{\{0-4\},\{5-9\},\ldots, \{95-99\}\}$. Note, following GEM, all the algorithms are evaluated in multi-head setting
(refer Sec.~\red{2.1} of the main paper).
We refer GEM~\cite{lopez2017gradient} to report the accuracies of iCaRL and GEM.
From the Tab.~\ref{tab:resnet_comp}, it can be seen that {\em RWalk outperforms both the methods by a significant margin}.

\subsection{Effect of Regularization Hyperparameter ($\lambda$) }
\label{sec:lambda_exper}
In Tab.~\ref{tab:lamda_insensitivity} we analyse the sensitivity of different methods to the regularization hyperparameter ($\lambda$). As evident, RWalk is less sensitive to $\lambda$ compared to {\ewc}
\footnote{By {\ewc} we always mean its faster version {\ewcp}.}~\cite{Kirkpatrick2016EWC} and PI~\cite{Zenke2017Continual}. This is because of the normalization of the Fisher and Path-based importance scores in RWalk. For example, as we vary $\lambda$ by a factor of $1\times10^5$ on MNIST, the {\em forgetting} and {\em intransigence} measures changed by $-0.06$ and $0.14$ on {\ewc}~\cite{Kirkpatrick2016EWC}, and $-0.07$ and $0.13$ on {\pii}~\cite{Zenke2017Continual}, respectively. On the other hand, the change in RWalk, as can be seen in the Tab.~\ref{tab:lamda_insensitivity}, is $0$ for both the measures. On CIFAR-100 a similar trend is observed in Tab.~\ref{tab:lamda_insensitivity}.

\begin{table}[t]
\centering
\caption{\em Comparison of different methods on MNIST and CIFAR-100 as the regularization strength ($\lambda$) is varied. With Forgetting and Intransigence we also provide the change ($\Delta$) in the corresponding measures, where the first row in each method is taken as the reference. As discussed in 
Sec.~\red{4.1} 
in the main paper, RWalk is less sensitive to $\lambda$ compared to {\ewc} and {\pii}, making it more appealing for  incremental learning.}
\label{tab:lamda_insensitivity}
\begin{tabular}{@{}c@{\hspace{2pt}}c@{\hspace{1pt}}c@{\hspace{1pt}}c@{\hspace{4pt}}c@{\hspace{4pt}}c@{\hspace{2pt}}c@{\hspace{0.5pt}}c@{\hspace{4pt}}c@{}}
\toprule
\multicolumn{1}{c}{\textbf{Methods}} &\multicolumn{4}{c}{\textbf{MNIST}} &\multicolumn{4}{c}{\textbf{CIFAR}} \\
\midrule
                                     & $\lambda$  & $A_5$(\%)    & $F_5 (\Delta)$   & $I_5 (\Delta)$ & $\lambda$       & $A_{10}$(\%)      & $F_{10} (\Delta)$ & $I_{10} (\Delta)$ \\

\cmidrule(r){2-5} \cmidrule(l){6-9}
\multirow{3}{*}{{\ewc}} & 75              & 80.3 & 0.19 (0)     & 0.1 (0)     & 3.0           & 28.9 & 0.38 (0) & -0.17 (0)     \\
                                               & $75\times10^3$  & 79.2 & 0.15 (-0.04) & 0.21 (0.11) & 300           & 34.1 & 0.28 (-0.1) & -0.07 (0.1)     \\
                                               & $75\times10^5$  & 79.1 & 0.13 (-0.06) & 0.24 (0.14) & $3\times10^5$ & 33.7 & 0.27 (-0.11)& -0.03 (0.14)   \\
\cmidrule(r){2-9}                                              

\multirow{3}{*}{{\pii}} & 0.1              & 79.3  & 0.23 (0)      & 0.05 (0)    & 0.1           & 34.7 & 0.27 (0) & -0.07 (0)     \\
                                               & 100             & 80.3  & 0.15 (-0.08)  & 0.22 (0.17) & 10            & 34.3 & 0.26 (-0.01) & -0.04   (0.03)  \\
                                               & 10000           & 78.5  & 0.16 (-0.07)  & 0.18 (0.13) & $1\times10^4$ & 33.7 & 0.27 (0) & -0.06 (0.01)     \\
\cmidrule(r){2-9}

\multirow{3}{*}{\textbf{RWalk (Ours)}}         & 0.1             & 82.6  & 0.16 (0)    & 0.12 (0)   & 0.1           & 34.5 & 0.28 (0) & -0.06 (0)     \\
                                               & 100             & 81.6  & 0.16 (0)   & 0.14 (0.02) & 10            & 33.2 & 0.28 (0) & -0.06 (0)    \\
                                               & 10000           & 81.6  & 0.16 (0)   & 0.12 (0)    & $1\times10^4$ & 34.2 & 0.28 (0) & -0.05 (0.01)    \\
\bottomrule
\end{tabular}
\vspace{-0.5cm}
\end{table}

\subsection{CIFAR Architecture and Task-Level Analysis}
\label{sec:supp_network_arch}
\begin{sidewaysfigure}[t]
	\begin{subfigure}{0.8\linewidth}
		\includegraphics[scale=0.27]{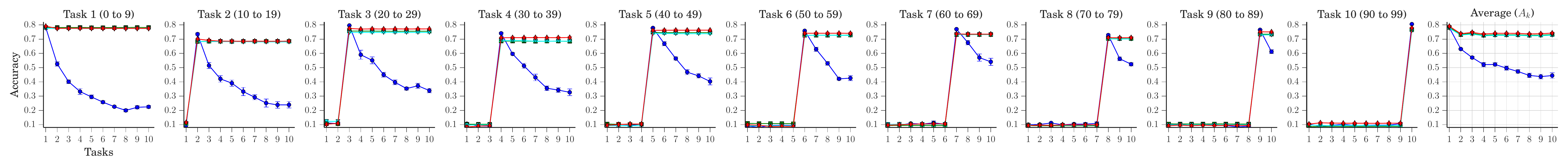}
	\end{subfigure}%

    \begin{subfigure}{0.8\linewidth}
		\includegraphics[scale=0.27]{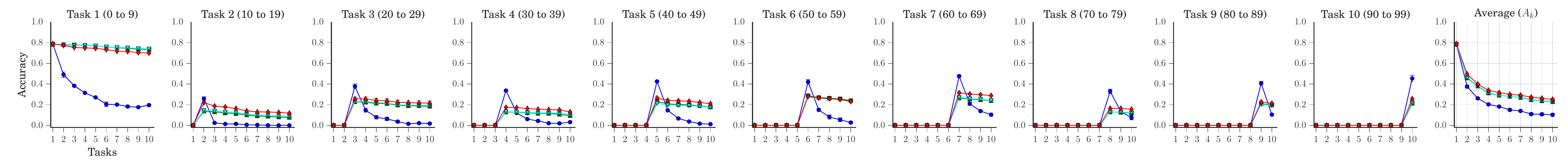}
	\end{subfigure}%
	
	\begin{subfigure}{0.8\linewidth}
		\includegraphics[scale=0.27]{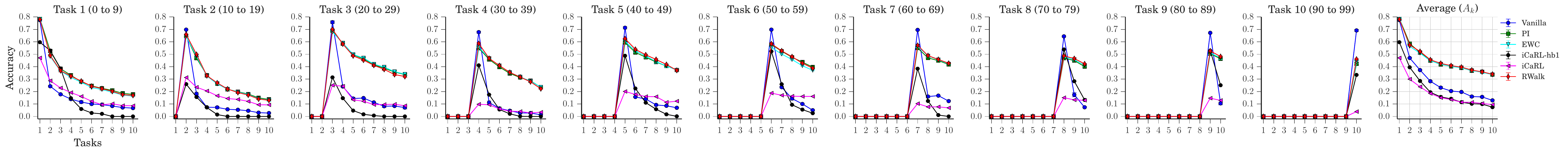}
	\end{subfigure}
	\caption{\em Accuracy measure in incremental CIFAR-100 with multi-head evaluation
(\textbf{top}), and the single-head evaluation without (\textbf{middle})
and with samples (\textbf{bottom}). The first ten columns show how the
performance of different tasks vary as the model is trained for new tasks, \eg,
the first plot depicts the variation in performance on Task 1 when the network
is sequentially trained for the ten tasks in an incremental manner. The last
column shows the average accuracy measure ($A_k$, by varying $k$). 
Mean of features (MoF) sampling is used. (best
viewed in color)}
   \label{fig:cifar_avg_acc}
\end{sidewaysfigure}

In Tab.~\ref{tab:cifar_conv_arch} we report the detailed architecture of the convolutional network used in the incremental CIFAR-100 experiments 
(Sec.~\red{6}). 
Note that, in contrast to {\pii}~\cite{Zenke2017Continual}, we use only one fully-connected layer (denoted as `FC' in the table). For each task $k$, the weights in the last layer of the network is dynamically added. Additionally, in Fig~\ref{fig:cifar_avg_acc}, we present a similar task-level analysis on CIFAR-100 as done for MNIST (Fig.~\red{2} in the main paper). Note that, for all the experiments `$\alpha$' in 
Eq.~(\red{6})
is set to $0.9$ and  `$\Delta t$' in 
Eq.~(\red{7})
is $10$ and $50$ for MNIST and CIFAR, respectively.

\begin{table}[h]
\centering
\caption{\em CNN architecture for incremental CIFAR-100 used for Vanilla, {\ewc}, {\pii}, iCaRl, RWalk in the main paper. Here, `$n$' denotes the number of classes in each task.}
\label{tab:cifar_conv_arch}
\setlength\tabcolsep{3pt}
\begin{tabular}{c@{\hspace{10pt}}c@{\hspace{6pt}}c@{\hspace{6pt}}c@{\hspace{6pt}}c@{\hspace{6pt}}c}
\toprule
\textbf{Operation}     & \textbf{Kernel} 		   & \textbf{Stride} 			 & \textbf{Filters} 	& \textbf{Dropout} & \textbf{Nonlin}. \\ \hline
3x32x32 input &        		   &        &         &         	&         \\
Conv          & $3 \times 3$    & $ 1 \times 1$    & $32$     &         & ReLU    \\
Conv          & $3\times 3$     & $1 \times 1$     & $32$     &         & ReLU    \\
MaxPool       &        		   & $2 \times 2$     &         	 & $0.5$   &         \\
Conv          & $3\times 3$    & $1 \times 1$     & $64$      &         & ReLU    \\
Conv          & $3\times 3$    & $1 \times 1$     & $64$      &         & ReLU    \\
MaxPool       &        		  & $2\times2$       &            & $0.5$   &         \\ \cmidrule{2-6}
Task 1: FC    &        		  &        			 & $n$        &         &         \\ \cmidrule{2-6}
$\cdots$: FC  &        		  &       			 & $n$        &         &         \\ \cmidrule{2-6}
Task k: FC    &        		  &        			 & $n$        &         &         \\
\bottomrule
\end{tabular}
\end{table}

\clearpage


\bibliographystyle{splncs04}
\bibliography{dokaniaBibliography}
\end{document}